\documentclass{article} 
\usepackage{iclr2027_conference,times}

\usepackage{amsmath,amsfonts,bm}

\def\eqref#1{equation~\ref{#1}}

\def\1{\bm{1}}

\DeclareMathAlphabet{\mathsfit}{\encodingdefault}{\sfdefault}{m}{sl}
\SetMathAlphabet{\mathsfit}{bold}{\encodingdefault}{\sfdefault}{bx}{n}

\usepackage{hyperref}
\usepackage{url}

\title{BARTER: Compact Dynamic Sparse Training with Standard Dense Kernels}

\author{
Yueyuan Sui$^{1}$ \quad
Payal Mohapatra$^{2}$ \quad
Qi Zhu$^{1}$ \quad
Stephen Xia$^{1}$ \\
[0.5em]
\normalfont\normalsize
$^{1}$Northwestern University, USA\qquad
$^{2}$Arm Inc., USA 
}

\hypersetup{hidelinks}
\newcommand{\method}{\texttt{BARTER}}

\usepackage{booktabs}
\usepackage{multirow}
\usepackage{xcolor}

\usepackage{graphicx}
\usepackage{wrapfig}
\usepackage{caption}
\usepackage{bm}

\usepackage{amsmath,amssymb,booktabs}
\usepackage{algorithm}
\usepackage{algpseudocode}

\definecolor{barterorange}{RGB}{166,74,0}

\newcommand{\barterscore}[2]{%
  #1\textsubscript{\ensuremath{\pm #2}}}

\newcommand{\barterbest}[2]{%
  \textcolor{barterorange}{\textbf{#1}}%
  \textsubscript{\ensuremath{\pm #2}}}

\newcommand{\bartersecond}[2]{%
  \underline{\textit{#1}}%
  \textsubscript{\ensuremath{\pm #2}}}

\iclrfinalcopy 
\begin{document}

\maketitle
\fancyhead[L]{Preprint}

\begin{abstract}
Dynamic sparse training (DST) aims to reduce training costs while maintaining accuracy. Existing DST methods mask out parameters but keep matrix dimensions unchanged. Training speedups therefore depend on specialized support, such as sparse kernels. We introduce \method{}, a structured DST framework that reduces training cost using standard dense kernels by keeping training tensors physically compact while dynamically reallocating capacity across layers. Unlike existing works that require computing gradients for all parameters, \method{} instead uses statistics collected only from active units during ordinary backpropagation to prune and clone attention heads and feed-forward channels under separate fixed global budgets. At each topology update, it resizes the affected parameter tensors and remaps their optimizer states, so subsequent training steps operate directly on the compact model. Training and fine-tuning experiments with Transformer-based vision models on two image classification benchmarks at densities from $0.1$ to $0.7$ show competitive Top-1 accuracy relative to full-tensor DST baselines. On an NVIDIA L40, \method{} achieves speedups of up to $6.29\times$ in measured training segments and reduces peak allocated GPU memory by up to $81.3\%$ relative to these implementations. An iPhone 13 Pro Max training benchmark additionally shows a $2.78\times$ speedup over the full-tensor DST baseline.
\end{abstract}

\section{Introduction}
\label{sec:introduction}

\begin{wrapfigure}{r}{0.50\textwidth}
    \vspace{-\intextsep}
    \centering
    \captionsetup{font=small,skip=4pt}
    \includegraphics[width=\linewidth]{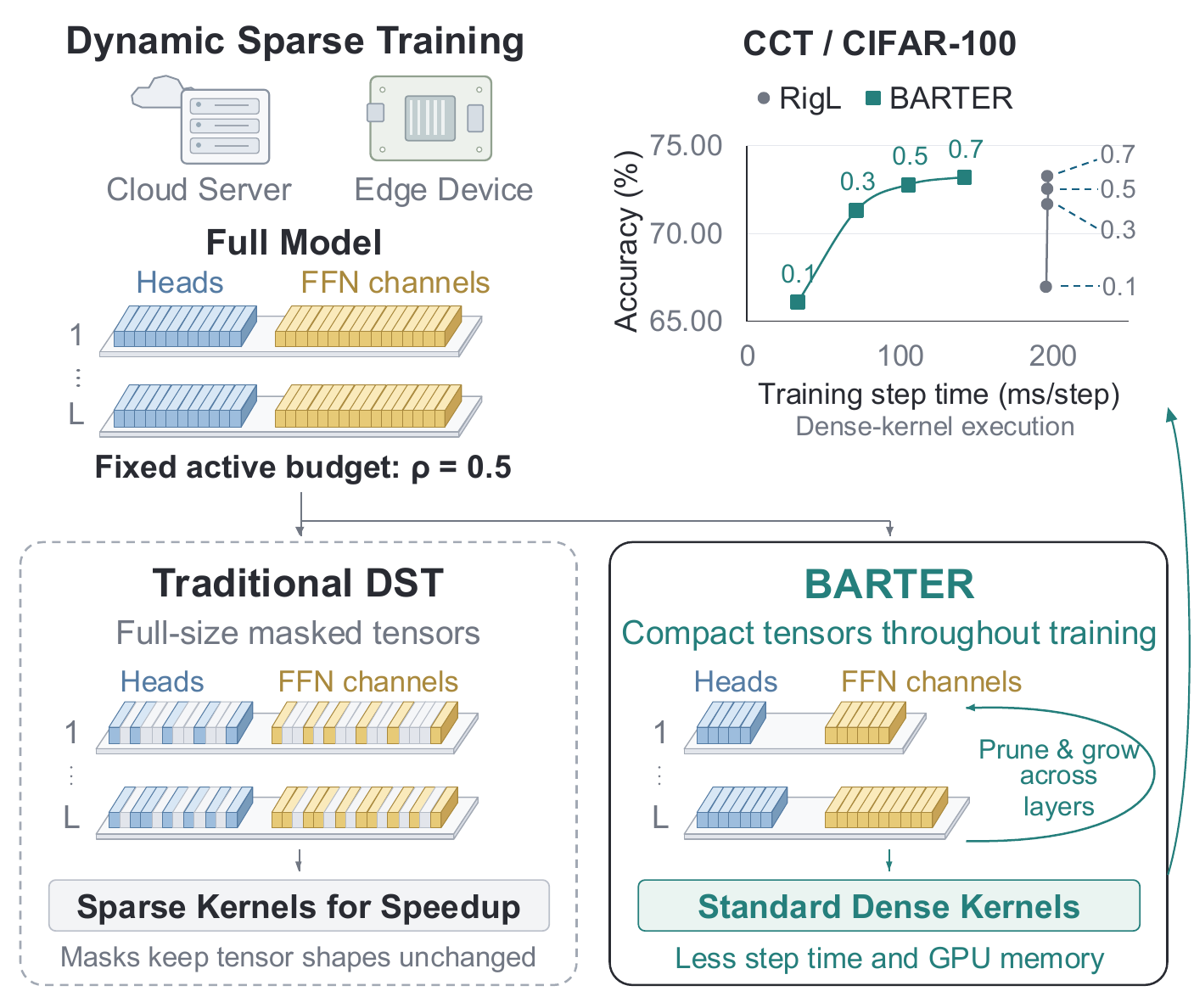}
    \caption{Full-size masked training versus compact dynamic training with \method{}. The plot compares CCT/CIFAR-100 accuracy and schedule-weighted training step-time estimates for RigL and \method{} under dense-kernel execution. Labels indicate density.}
    \label{fig:barter_teaser}
    \vspace{\dimexpr-\intextsep-10pt\relax}
\end{wrapfigure}

Training neural networks requires substantial computation and memory, while deployment imposes limits on model size and inference cost. Pruning reduces deployment costs by removing redundant weights~\citep{han2015learning} or structural units~\citep{li2017pruning}, but training a dense model before pruning still incurs the full training cost. Dynamic sparse training (DST)~\citep{set,snfs,rigl} aims to reduce this training cost while maintaining accuracy. It trains a sparse network from initialization under a limited active-parameter budget. The initial topology may not make the best use of this budget, so DST periodically prunes and grows connections to revise its capacity allocation as it learns.

However, fewer active parameters do not necessarily mean cheaper training. Existing DST methods mask out parameters but keep matrix dimensions unchanged~\citep{rigl,s2vite,srigl}. Standard dense kernels still process the full-size tensors. Sparse kernels can skip computations on inactive weights to accelerate training, but require implementations tailored to the sparsity format and execution backend~\citep{sparseprop,hu2024accelerating,slope}. This limits which platforms can benefit from these speedups.

To achieve reductions in computation with standard dense kernels, a logical approach is to remove inactive structural units and train on a smaller tensor consisting only of active units. However, it is not possible to do so using existing approaches because they require using the gradients of inactive candidates to guide growth~\citep{rigl,s2vite,srigl}; once these units are physically removed, ordinary backpropagation no longer provides those gradients. 

\noindent\textbf{Our Approach.}
We propose \method{}, a structured DST framework that prunes and grows structural units directly on compact tensors using standard dense kernels (Figure~\ref{fig:barter_teaser}). Under fixed global \underline{B}udgets, \method{} uses only \underline{A}ctive-unit statistics to \underline{R}eallocate capacity during \underline{T}raining. It maintains compact dense \underline{E}xecution by \underline{R}esizing the affected tensors. In our Transformer implementation, attention heads and feed-forward network (FFN) channels have separate budgets, and training uses compact tensors from the first optimization step. At each update, \method{} uses these statistics to prune a unit from a donor layer and clone an active unit within the receiving layer. The clone starts from a feature the model has already learned, and the parent's outgoing weights are split between the two units. We resize the affected parameters and remap their optimizer states so training can continue on the compact model. The final model can be deployed without further pruning.

Our contributions are summarized as follows:
\begin{itemize}
    \item We introduce \method{}, a structured DST framework that reallocates capacity under fixed global budgets while keeping training tensors physically compact. Unlike existing DST implementations that rely on specialized sparse kernels for acceleration, \method{} achieves training speedups using standard dense kernels and produces a compact model for deployment.

    \item We develop a reallocation mechanism that selects units for pruning and cloning. Unlike prior growth rules that rely on gradients of inactive candidates, it uses only active-unit statistics. Tensor resizing and optimizer-state remapping allow training to continue directly on the compact model after each structural update.

    \item We conduct training and fine-tuning experiments with two Transformer architectures on two image classification benchmarks, comparing against five state-of-the-art DST methods and a dense reference. Under standard dense-kernel execution, \method{} maintains competitive classification performance while achieving speedups of up to $6.29\times$ in measured GPU training segments and reducing peak allocated GPU memory by up to $81.3\%$ relative to these DST methods. A smartphone training benchmark shows a $2.78\times$ speedup over RigL.
\end{itemize}

\section{Related Work}
\label{sec:related_work}

\paragraph{Unstructured dynamic sparse training.} Unstructured DST learns sparse connectivity alongside network weights under limited connection budgets~\citep{deepr,set}. SET~\citep{set} combines magnitude-based pruning with random growth, while RigL~\citep{rigl} selects new connections using gradient magnitudes. CHTs~\citep{chts} uses topology-based link prediction to guide regrowth. Beyond the selection of individual connections, dynamic sparse reparameterization redistributes a fixed nonzero parameter budget across layers~\citep{dsr}, and in-time over-parameterization relates training performance to the exploration of sparse configurations over time~\citep{itop}. These methods adapt connectivity at the level of individual weights. \method{} changes the granularity of reallocation to complete attention heads and FFN channels, physically adjusting the corresponding tensor dimensions as capacity moves across layers.

\paragraph{Structured dynamic sparse training.} Structured DST introduces constraints on the units or connectivity patterns being updated. S$^2$ViTE~\citep{s2vite} prunes and grows attention heads and FFN channels; Chase~\citep{chase} combines channel pruning with weight-level redistribution; and SRigL~\citep{srigl} maintains constant fan-in connectivity with neuron ablation. Their growth rules use gradients of inactive candidate weights or units~\citep{s2vite,chase,srigl}, which ordinary backpropagation through a physically compact active model does not provide. \method{} supports growth by cloning an existing active unit within the receiving layer, with reallocation decisions based only on active-unit statistics. Transfers preserve separate global budgets for attention heads and FFN channels, while parameter resizing and optimizer-state remapping keep subsequent training compact.

\paragraph{Sparse execution for training.} MEST~\citep{mest} combines memory-economic sparse training with compiler-level optimizations for mobile execution. DSB~\citep{dsb} organizes changing connectivity into shuffled blocks for efficient sparse computation, while DynaDiag~\citep{dynadiag} maintains diagonal sparsity and accelerates training with custom CUDA kernels. For large output spaces, HASTE~\citep{haste} pairs group-shared fixed fan-in patterns with custom CUDA kernels. Transformer pre-training has also benefited from optimized forward and backward kernels for 2:4 sparsity~\citep{hu2024accelerating}, while CHTs24~\citep{chts24} studies dynamic connectivity updates under transposable 2:4 constraints. These approaches tailor execution to particular sparsity patterns and backends. \method{} follows a different route: it reduces layer dimensions during training and executes the resulting compact tensors with standard dense kernels.
\section{Methodology}
\label{sec:method}

\begin{figure}[t]
    \centering
    \includegraphics[width=\linewidth]{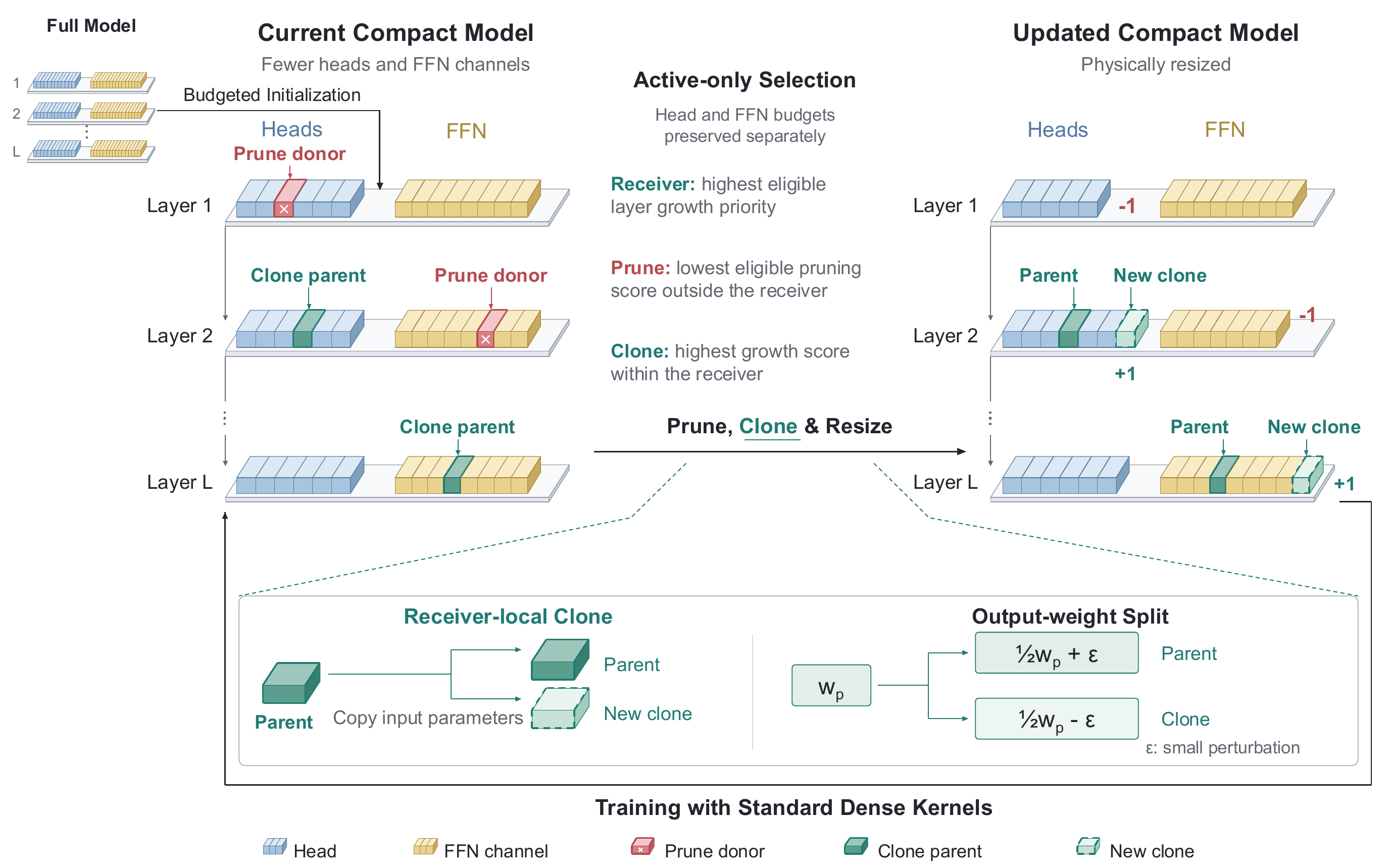}
    \caption{Overview of our approach, \method{}. Budgeted initialization produces a compact model with fewer attention heads and FFN channels while retaining all Transformer layers. During training, statistics from active units guide capacity reallocation across layers. For each unit type, the method selects the eligible receiving layer with the highest growth priority, prunes eligible units with the lowest pruning scores outside that layer, and adds the same number of units by cloning parents with the highest growth scores within it. These updates preserve the global head and FFN budgets separately. Each clone copies its parent's input parameters, while the parent's output weights are split between the parent and clone. The affected parameter tensors and optimizer states are resized accordingly, and training continues on the compact model with standard dense kernels.}
    \label{fig:barter_overview}
    \vspace{-\intextsep}
\end{figure}

\method{} adapts the allocation of attention heads and feed-forward
channels across layers while training a physically compact model.
Figure~\ref{fig:barter_overview} provides an overview of the framework.
We first define the compact model and its capacity budgets
(Section~\ref{sec:capacity_budgets}), then describe how statistics
from active units guide cross-layer reallocation
(Section~\ref{sec:active_reallocation}).
We next present the prune-and-clone operation
(Section~\ref{sec:prune_clone}) and explain how the affected
parameters and optimizer states are resized so that training
can continue with standard dense kernels
(Section~\ref{sec:physical_resizing}).

\subsection{Compact Model and Capacity Budgets}
\label{sec:capacity_budgets}

Consider a model with $L$ Transformer layers.
We denote the numbers of active attention heads and
feed-forward channels in layer $\ell$ by $h_\ell$
and $c_\ell$, respectively.
We store the parameters of these active units in compact
dense tensors and train the model with standard dense kernels.

At initialization, the target density $\rho$ specifies
the fraction of the original model's prunable weights
to retain.
We denote the global budgets for attention heads and
feed-forward channels by $H$ and $C$, respectively:
\begin{equation}
    H = \sum_{\ell=1}^{L} h_\ell,
    \qquad
    C = \sum_{\ell=1}^{L} c_\ell.
    \label{eq:capacity_budgets}
\end{equation}
We choose integer values for these budgets based on $\rho$
and the parameter costs of the two unit types, keeping
the retained fractions of heads and channels close to $\rho$.
We distribute each budget as evenly as possible across
layers, assigning any remainder to randomly chosen layers.
Within each layer of the original model, we randomly
select the allocated number of units and copy their
parameters into the initial compact model.

Training begins with this compact model.
During training, \method{} uses statistics from active
units to guide capacity reallocation across layers.
Each transfer removes a unit from one layer and adds
a unit of the same type to another, keeping the global
budgets in Eq.~(\ref{eq:capacity_budgets}) fixed.

\subsection{Active-Only Capacity Reallocation}
\label{sec:active_reallocation}

\paragraph{Active-unit scoring.}
We use activations and their gradients from ordinary training
to decide where to remove and add capacity.
For attention heads, we collect the head outputs before
the output projection; for FFN channels, we collect the
hidden activations after GELU.

For an active unit $i$ in layer $\ell$, let
$a_{\ell,i,b,j}$ denote its activation for sample $b$
at output coordinate $j$.
Here, $b$ indexes samples in the current minibatch.
For an FFN channel, $j$ runs over token positions;
for an attention head, it runs over token positions
and features within the head.
Let $\mathcal{L}$ be the minibatch loss and
$g_{\ell,i,b,j}
= \partial\mathcal{L}/\partial a_{\ell,i,b,j}$
the gradient with respect to that activation.

We multiply each activation value by its corresponding
loss gradient~\citep{molchanov2016pruning,michel2019sixteen}.
We aggregate these products into two minibatch scores:
$d_{\ell i}$ for pruning and $s_{\ell i}$ for growth.
Because pruning removes all outputs of the unit,
we compute $d_{\ell i}$ by summing the signed
activation-gradient products over all output coordinates
within each sample, taking the absolute value, and then
averaging across samples.
For $s_{\ell i}$, we take the absolute value of each
product before averaging over coordinates and samples,
so contributions of opposite signs do not cancel:
\begin{equation}
\begin{aligned}
d_{\ell i}
&=
\operatorname{mean}_{b}
\left|
\sum_j a_{\ell,i,b,j}g_{\ell,i,b,j}
\right|,\\
s_{\ell i}
&=
\operatorname{mean}_{b}
\left[
\operatorname{mean}_{j}
\left|a_{\ell,i,b,j}g_{\ell,i,b,j}\right|
\right].
\end{aligned}
\label{eq:batch_unit_scores}
\end{equation}

We maintain running pruning and growth scores,
$D_{\ell i}$ and $S_{\ell i}$, using exponential moving
averages (EMAs) of $d_{\ell i}$ and
$s_{\ell i}$~\citep{molchanov2019importance}.
The decay factor $\beta$ determines how much of
the previous running score is retained:
\begin{equation}
\begin{aligned}
D_{\ell i}
&\leftarrow \beta D_{\ell i}+(1-\beta)d_{\ell i},\\
S_{\ell i}
&\leftarrow \beta S_{\ell i}+(1-\beta)s_{\ell i}.
\end{aligned}
\label{eq:active_unit_scores}
\end{equation}
Both EMAs are initialized to zero at the start of training
and updated during each ordinary backward pass while
score collection is enabled.
At scheduled topology updates, these running scores
guide the selections below.

\paragraph{Receiver selection.}
We apply the selection procedure separately to attention heads and FFN channels. For the type being transferred, let $\mathcal{U}_\ell$ be the set of active unit indices in layer $\ell$. Its current unit count is $n_\ell=|\mathcal{U}_\ell|$: $n_\ell=h_\ell$ for attention heads and $n_\ell=c_\ell$ for FFN channels. We denote the count at compact initialization by $n_{\ell,0}$ and the count in the original model by $n_\ell^{\max}$.

A receiving layer must contain an active unit to clone and have room to grow, giving $0<n_\ell<n_\ell^{\max}$. For each eligible layer, we multiply the mean growth score of its active units by the ratio of its initial to current unit count. This gives the corrected layer score $R_\ell$. We select the eligible layer with the highest score as the receiver:
\begin{equation}
\begin{aligned}
R_\ell &= \frac{n_{\ell,0}}{n_\ell}\frac{1}{n_\ell}\sum_{i\in\mathcal{U}_\ell}S_{\ell i},\\
r &= \operatorname*{arg\,max}_{\ell:\,0<n_\ell<n_\ell^{\max}}R_\ell.
\end{aligned}
\label{eq:receiver_selection}
\end{equation}
The width correction lowers the priority of layers that have expanded relative to initialization and raises that of layers that have contracted.

\paragraph{Donor and clone-source selection.}
With the receiving layer $r$ fixed, let $\mathcal{C}_r$
contain the eligible donor pairs $(\ell,i)$ outside
layer $r$, where $\ell$ is the layer index and $i$
is the unit index within that layer.
Removing a unit must leave at least one FFN channel
per layer, while attention layers may have zero heads.
Recently cloned parents and their new copies are
excluded from donor selection during a cooldown period.

We choose the eligible unit with the lowest running
pruning score $D_{\ell i}$.
Its layer and within-layer indices are denoted by
$\ell_d$ and $i_d$, respectively.
Within the receiving layer, we choose the active unit
with the highest running growth score $S_{ri}$
as the clone parent and denote its index by $i_p$:
\begin{equation}
\begin{aligned}
(\ell_d,i_d)
&=
\operatorname*{arg\,min}_{(\ell,i)\in\mathcal{C}_r}
D_{\ell i},\\
i_p
&=
\operatorname*{arg\,max}_{i\in\mathcal{U}_r}
S_{ri}.
\end{aligned}
\label{eq:donor_parent_selection}
\end{equation}
We skip a transfer if no receiving layer is eligible
or the selected receiver has no eligible donor.
Otherwise, layer $\ell_d$ releases unit $i_d$,
and layer $r$ gains a unit of the same type
by cloning its own active unit $i_p$.
Section~\ref{sec:prune_clone} describes this update.

\subsection{Prune-and-Clone Updates}
\label{sec:prune_clone}

\paragraph{Unit removal and cloning.}
Using the selections from Section~\ref{sec:active_reallocation},
\method{} removes the donor unit $(\ell_d,i_d)$ and
creates a unit of the same type in receiving layer $r$
by cloning its selected parent $i_p$.
We remove an attention head's query, key, and value
parameter blocks, including their biases when present,
together with the corresponding output-projection weights.
For an FFN channel, we remove its input weights,
bias, and output weights.
The new unit, indexed by $i_{\mathrm{new}}$ within
layer $r$, is initialized from its
parent~\citep{wu2019splitting,wu2020firefly}.
We copy the parent's parameters for producing features
and initialize the two units' output weights jointly
through the split below.

\paragraph{Output-weight splitting.}
We divide the parent's output weights between the parent
and the new unit to avoid duplicating its original
output contribution~\citep{chen2015net2net}.
Let $W^{\mathrm{out}}_{r,i}$ denote the weights that map
unit $i$'s features to the output of its module in layer $r$,
and let $W^{\mathrm{out,old}}_{r,i_p}$ denote the parent's
output weights before cloning.
Using a small Gaussian perturbation $\varepsilon$
with the same shape as these weights, we set
\begin{equation}
\begin{aligned}
W^{\mathrm{out}}_{r,i_p}
&\leftarrow
\tfrac{1}{2}W^{\mathrm{out,old}}_{r,i_p}
+\varepsilon,\\
W^{\mathrm{out}}_{r,i_{\mathrm{new}}}
&\leftarrow
\tfrac{1}{2}W^{\mathrm{out,old}}_{r,i_p}
-\varepsilon.
\end{aligned}
\label{eq:clone_split}
\end{equation}
The opposite perturbations break the symmetry between
the two units while keeping the sum of their output
weights equal to the parent's original weights.
When the parent and its copy produce identical features,
their combined output contribution therefore equals
the parent's contribution before the split.

\paragraph{Score updates after cloning.}
We remove the donor's scores and retain the scores
of units unaffected by the update.
Let $D^{\mathrm{old}}_{r,i_p}$ and
$S^{\mathrm{old}}_{r,i_p}$ denote the parent's running
pruning and growth scores before cloning.
We split each stored score equally between the parent
and its new copy, keeping their combined value unchanged:
\begin{equation}
\begin{aligned}
D_{r,i_p}
&\leftarrow \tfrac{1}{2}D^{\mathrm{old}}_{r,i_p},
&
D_{r,i_{\mathrm{new}}}
&\leftarrow \tfrac{1}{2}D^{\mathrm{old}}_{r,i_p},\\
S_{r,i_p}
&\leftarrow \tfrac{1}{2}S^{\mathrm{old}}_{r,i_p},
&
S_{r,i_{\mathrm{new}}}
&\leftarrow \tfrac{1}{2}S^{\mathrm{old}}_{r,i_p}.
\end{aligned}
\label{eq:clone_score_initialization}
\end{equation}
Both units enter the donor-selection cooldown defined
in Section~\ref{sec:active_reallocation}.
Subsequent training updates their scores from their
own activations and gradients using
Equation~\ref{eq:active_unit_scores}.
Section~\ref{sec:physical_resizing} describes how
the affected parameter tensors and optimizer states
are resized for continued training.

\subsection{Physical Resizing and Continued Training}
\label{sec:physical_resizing}

\paragraph{Physical resizing.}
At each topology update, we rebuild the affected tensors
from the retained and newly created parameter slices
defined in Section~\ref{sec:prune_clone}.
We use matching unit indices across the associated tensors.
For an FFN with $c_\ell$ active channels and embedding
dimension $d$, the input and output weight matrices
have sizes $c_\ell \times d$ and $d \times c_\ell$,
respectively.
Attention projections are resized by complete head blocks.
Both modules retain their input and output dimensions
as their internal widths change.
Removed units leave no zero-filled slots, and subsequent
forward and backward passes operate directly on the
resized tensors with standard dense kernels.

\paragraph{Optimizer-state remapping.}
We resize the optimizer states using the same unit indices
as the parameters.
For Adam, we retain the first- and second-moment estimates
of unaffected parameter slices and remove those belonging
to the pruned unit.
Both estimates are set to zero for the parameters of
the cloned parent and its new copy, while the Adam
step counters remain unchanged.

\paragraph{Training schedule.}
Pruning decisions can benefit from training beyond
initialization~\citep{frankle2020pruning}.
We therefore train the initial compact model and accumulate
active-unit scores before the first capacity transfer.
Let $T$ denote the total number of training steps,
and let $t_{\mathrm{start}}$ and $t_{\mathrm{end}}$
define the reallocation window.
We set the total numbers of head and FFN-channel
transfer attempts, $K_H$ and $K_C$, in proportion
to their initial active counts $H$ and $C$, respectively.
For each unit type, we spread these attempts approximately
uniformly across the window.
Training between updates lets the model adapt to its
current structure and refresh the importance estimates
used for the next decision~\citep{s2vite}.
At each scheduled update, we attempt one transfer
of the corresponding unit type after the optimizer step,
keeping the change small to limit disruption
to training~\citep{chase}.

We stop collecting scores for each unit type after
its last scheduled update.
After $t_{\mathrm{end}}$, we fix the layer widths
and continue optimizing the weights until step $T$,
as sparse connectivity can stabilize earlier
than the weights~\citep{dsr}. Appendix~\ref{app:reallocation_sensitivity} examines how increasing the number of capacity transfers affects classification performance and incremental update cost.
\section{Results and Discussion}
\label{sec:results}

We evaluate \method{} against dense and sparse baselines on two image classification benchmarks, comparing classification performance, training cost, and final model size.

\paragraph{Datasets and models.} Our from-scratch experiments use a seven-layer CCT~\citep{cct} on CIFAR-100~\citep{cifar100} and a nine-layer ViT~\citep{vit} on Tiny-ImageNet~\citep{tinyimagenet}. CCT uses $32\times32$ inputs, while ViT uses $64\times64$ inputs and the small-model configuration of~\citet{gani2022train}. We also fine-tune the Tiny-ImageNet-trained ViT on CIFAR-100 with images resized to $64\times64$. We reserve $10\%$ of each official training set for validation. Final evaluation uses the CIFAR-100 test set and the labeled Tiny-ImageNet validation set.

\paragraph{Baselines.} We compare \method{} with Dense, the unstructured DST methods SET~\citep{set}, SNFS~\citep{snfs}, and RigL~\citep{rigl}, and the structured DST methods SRigL~\citep{srigl} and S$^2$ViTE~\citep{s2vite}.We use the official implementations of the DST baselines, adapted to our model configurations. Sparse models use target densities $\rho\in\{0.1,0.3,0.5,0.7\}$. Initial capacity allocation and baseline settings are described in Appendices~\ref{app:datasets_models} and~\ref{app:baseline_ablation_settings}.

\paragraph{Training settings.} In the fixed-step comparisons, methods share the data split, augmentation, and optimization settings within each task and seed. GPU training uses FP32 and Adam~\citep{adam} with decoupled weight decay~\citep{adamw}. CCT is trained for 200 epochs. ViT training and standard fine-tuning each run for 100 epochs, with three shuffled passes through the training set per epoch. For \method{}, we set the score EMA decay to $0.99$ and reallocate capacity between $10\%$ and $40\%$ of the training steps. Detailed training settings and checkpoint-selection rules are provided in Appendix~\ref{app:training_settings}.

\paragraph{Evaluation.} We select checkpoints by validation Top-1 accuracy and report Top-1 accuracy and macro-F1 as mean $\pm$ sample standard deviation across three random seeds. For training efficiency, we report measured training time for epoch-length segments, estimated per-step time, and peak allocated GPU memory. We also report estimated training FLOPs, counting each multiply-accumulate operation as two FLOPs, and the physical parameter counts of the exported models. Measurement and counting protocols are provided in Appendix~\ref{app:efficiency}.

\subsection{Classification Performance}
\label{sec:classification}

\begin{table}[t]
\centering
\caption{Classification performance (accuracy/macro-F1) of models
trained from scratch. Best sparse results per task and density in
\textcolor{barterorange}{\textbf{bold and orange}}; second-best in
\underline{\textit{underlined italics}}.
Mean $\pm$ std (\%) over three seeds.}
\label{tab:classification}
\begingroup
\normalsize
\setlength{\tabcolsep}{1.5pt}
\renewcommand{\arraystretch}{1.12}
\resizebox{\textwidth}{!}{%
\begin{tabular}{@{}l*{8}{c}@{}}
\toprule
\multicolumn{9}{c}{\textbf{(a) CIFAR-100 (CCT)}} \\
\multicolumn{9}{c}{Dense ($\rho=1$): Acc \barterscore{73.49}{0.10},
\quad F1 \barterscore{73.29}{0.11}} \\
\addlinespace[2pt]
& \multicolumn{2}{c}{$\rho=0.1$}
  & \multicolumn{2}{c}{$\rho=0.3$}
  & \multicolumn{2}{c}{$\rho=0.5$}
  & \multicolumn{2}{c}{$\rho=0.7$} \\
\cmidrule(lr){2-3}\cmidrule(lr){4-5}
\cmidrule(lr){6-7}\cmidrule(lr){8-9}
Method
& Acc $\uparrow$ & F1 $\uparrow$ & Acc $\uparrow$ & F1 $\uparrow$
& Acc $\uparrow$ & F1 $\uparrow$ & Acc $\uparrow$ & F1 $\uparrow$ \\
\midrule
SET
& \barterscore{63.21}{0.55} & \barterscore{62.80}{0.56} & \barterscore{70.26}{0.30} & \barterscore{70.08}{0.31}
& \barterscore{72.25}{0.06} & \barterscore{72.04}{0.10} & \barterscore{73.21}{0.16} & \barterscore{72.98}{0.13} \\
SNFS
& \barterscore{63.29}{0.36} & \barterscore{62.87}{0.35} & \barterscore{69.45}{0.14} & \barterscore{69.31}{0.14}
& \barterscore{70.92}{0.53} & \barterscore{70.74}{0.51} & \barterscore{72.17}{0.17} & \barterscore{71.97}{0.12} \\
RigL
& \barterbest{67.01}{0.17} & \barterbest{66.67}{0.08} & \barterbest{71.74}{0.14} & \barterbest{71.57}{0.10}
& \bartersecond{72.62}{0.21} & \bartersecond{72.40}{0.22} & \barterbest{73.29}{0.08} & \barterbest{73.09}{0.07} \\
\addlinespace[2pt]
SRigL
& \barterscore{60.88}{0.82} & \barterscore{60.37}{0.85} & \barterscore{64.66}{0.44} & \barterscore{64.34}{0.42}
& \barterscore{69.94}{0.54} & \barterscore{69.79}{0.53} & \barterscore{72.40}{0.22} & \barterscore{72.17}{0.20} \\
S$^2$ViTE
& \bartersecond{66.64}{0.57} & \bartersecond{66.36}{0.60} & \barterscore{70.94}{0.27} & \barterscore{70.79}{0.27}
& \barterscore{72.49}{0.43} & \barterscore{72.27}{0.44} & \barterscore{73.10}{0.40} & \barterscore{72.88}{0.42} \\
\textsc{BARTER}
& \barterscore{66.13}{0.40} & \barterscore{65.79}{0.44} & \bartersecond{71.38}{0.36} & \bartersecond{71.21}{0.32}
& \barterbest{72.85}{0.16} & \barterbest{72.64}{0.14} & \bartersecond{73.26}{0.28} & \bartersecond{73.05}{0.28} \\
\midrule
\addlinespace[3pt]
\multicolumn{9}{c}{\textbf{(b) Tiny-ImageNet (ViT)}} \\
\multicolumn{9}{c}{Dense ($\rho=1$): Acc \barterscore{52.21}{0.29},
\quad F1 \barterscore{51.62}{0.33}} \\
\addlinespace[2pt]
& \multicolumn{2}{c}{$\rho=0.1$}
  & \multicolumn{2}{c}{$\rho=0.3$}
  & \multicolumn{2}{c}{$\rho=0.5$}
  & \multicolumn{2}{c}{$\rho=0.7$} \\
\cmidrule(lr){2-3}\cmidrule(lr){4-5}
\cmidrule(lr){6-7}\cmidrule(lr){8-9}
Method
& Acc $\uparrow$ & F1 $\uparrow$ & Acc $\uparrow$ & F1 $\uparrow$
& Acc $\uparrow$ & F1 $\uparrow$ & Acc $\uparrow$ & F1 $\uparrow$ \\
\midrule
SET
& \barterscore{43.44}{0.03} & \barterscore{42.34}{0.05} & \barterscore{51.22}{0.18} & \barterscore{50.73}{0.18}
& \bartersecond{52.39}{0.63} & \bartersecond{51.92}{0.72} & \barterscore{52.52}{0.38} & \barterscore{52.09}{0.36} \\
SNFS
& \bartersecond{43.72}{0.13} & \bartersecond{42.71}{0.18} & \barterbest{51.49}{0.30} & \barterbest{50.97}{0.33}
& \barterscore{52.16}{0.36} & \barterscore{51.74}{0.39} & \barterscore{52.12}{0.29} & \barterscore{51.72}{0.33} \\
RigL
& \barterbest{44.13}{0.52} & \barterbest{43.16}{0.51} & \barterscore{51.00}{0.14} & \barterscore{50.51}{0.14}
& \barterscore{51.76}{0.44} & \barterscore{51.32}{0.47} & \barterscore{51.58}{0.34} & \barterscore{51.10}{0.30} \\
\addlinespace[2pt]
SRigL
& \barterscore{42.06}{0.69} & \barterscore{40.86}{0.78} & \barterscore{44.29}{0.05} & \barterscore{43.28}{0.07}
& \barterscore{52.10}{0.34} & \barterscore{51.61}{0.32} & \barterbest{53.01}{0.61} & \bartersecond{52.59}{0.62} \\
S$^2$ViTE
& \barterscore{42.52}{0.83} & \barterscore{41.14}{0.91} & \barterscore{50.98}{0.32} & \barterscore{50.44}{0.30}
& \barterscore{52.16}{0.30} & \barterscore{51.72}{0.32} & \barterscore{52.68}{0.26} & \barterscore{52.23}{0.24} \\
\textsc{BARTER}
& \barterscore{42.56}{0.86} & \barterscore{41.14}{0.99} & \bartersecond{51.42}{0.39} & \bartersecond{50.81}{0.35}
& \barterbest{52.77}{0.23} & \barterbest{52.34}{0.29} & \bartersecond{53.00}{0.28} & \barterbest{52.64}{0.23} \\
\bottomrule
\end{tabular}%
}
\endgroup
\end{table}

Table~\ref{tab:classification} shows that \method{} ranks
first or second among sparse methods in both Top-1 accuracy
and macro-F1 at densities of $0.3$, $0.5$, and $0.7$
on CIFAR-100 and Tiny-ImageNet.
At $\rho=0.5$, it achieves the
\textbf{highest mean Top-1 accuracy and macro-F1}
among sparse methods on both tasks, with Top-1 accuracies
of $72.85\%$ and $52.77\%$, respectively.

Even at $\rho=0.1$, \method{}
\textbf{physically removes approximately $90\%$ of the
original attention heads and FFN channels},
while the baselines in our experiments retain full-size
training tensors. Despite this reduction, its Top-1 accuracy is only $0.88$ and $1.57$ percentage points below RigL, the best-performing sparse baseline at this density, on CIFAR-100 and Tiny-ImageNet, respectively.

\newpage

\subsection{Training Efficiency and Model Size}
\label{sec:training_efficiency}

\begin{figure}[t]
    \centering
    \includegraphics[width=\textwidth]
        {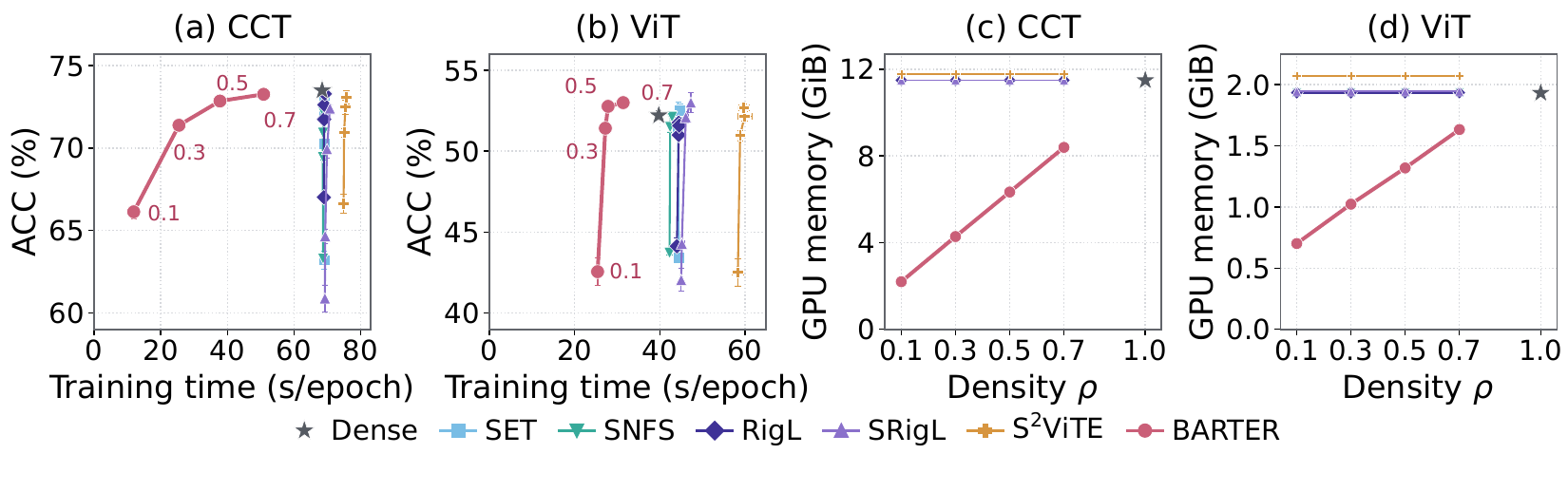}
    \caption{Accuracy, GPU training time per epoch, and peak
    allocated GPU memory for CCT on CIFAR-100 and ViT on
    Tiny-ImageNet. Colors distinguish methods.
    Each sparse-method curve connects four points corresponding
    to densities $\rho=0.1,0.3,0.5,0.7$.
    Dense is a single full-density reference ($\rho=1$).
    Error bars show sample standard deviations.}
    \label{fig:training_efficiency}
\end{figure}

\begin{wrapfigure}{r}{0.40\textwidth}
    \vspace{\dimexpr-\intextsep-10pt\relax}
    \centering
    \captionsetup{font=small,skip=4pt}
    \includegraphics[width=\linewidth]
        {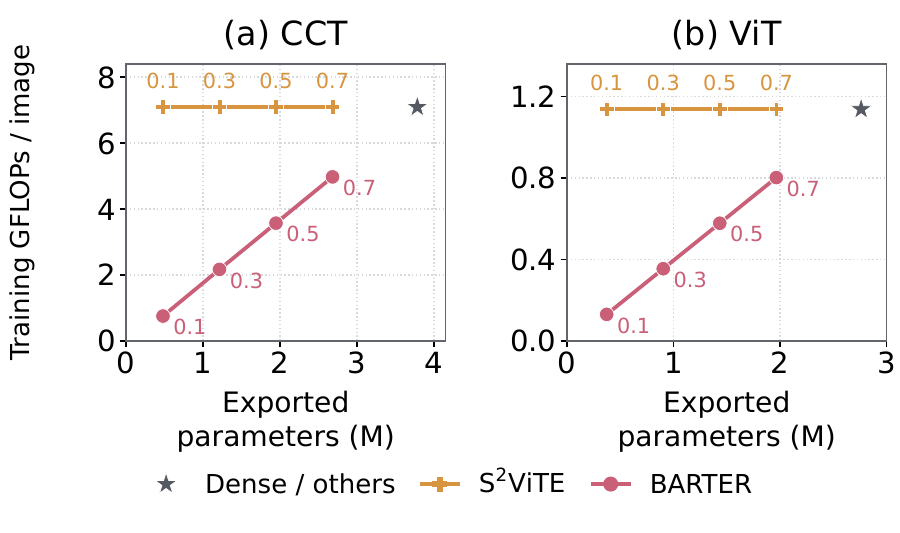}
    \caption{Estimated training GFLOPs per image versus exported parameter counts for (a) CCT and (b) ViT.
    Points on each colored curve correspond to densities
    $\rho=0.1,0.3,0.5,0.7$ from left to right.
    The star marks the overlapping results of Dense, SET,
    SNFS, RigL, and SRigL.
    S$^2$ViTE and \method{} have identical exported
    parameter counts at each density.}
    \label{fig:compute_model_size}
    \vspace{\dimexpr-\intextsep-10pt\relax}
\end{wrapfigure}

Figure~\ref{fig:training_efficiency} compares accuracy, GPU
training time per epoch, and peak allocated GPU memory. 
\textbf{\method{} has the lowest mean training time and peak allocated memory among all compared methods at every tested density.}
At $\rho=0.5$, it achieves the highest mean ACC and macro-F1
among sparse methods on both tasks, while training each epoch
$1.83\times$ faster than RigL on CCT and $1.60\times$ faster on ViT. The corresponding speedups over S$^2$ViTE are $2.00\times$ and $2.15\times$.


At $\rho=0.1$, speedups over RigL reach $5.80\times$ and
$1.73\times$, with peak allocated memory reduced by
$80.9\%$ and $63.8\%$, respectively.
Schedule-weighted step-time estimates show the same efficiency
advantage across all four densities
(Appendix~\ref{app:efficiency_results}).

Figure~\ref{fig:compute_model_size} compares estimated training FLOPs and exported parameter counts.

\begin{wrapfigure}{r}{0.4\textwidth}
    \vspace{\dimexpr-\intextsep-10pt\relax}
    \centering
    \captionsetup{font=small,skip=4pt}
    \includegraphics[width=\linewidth]
        {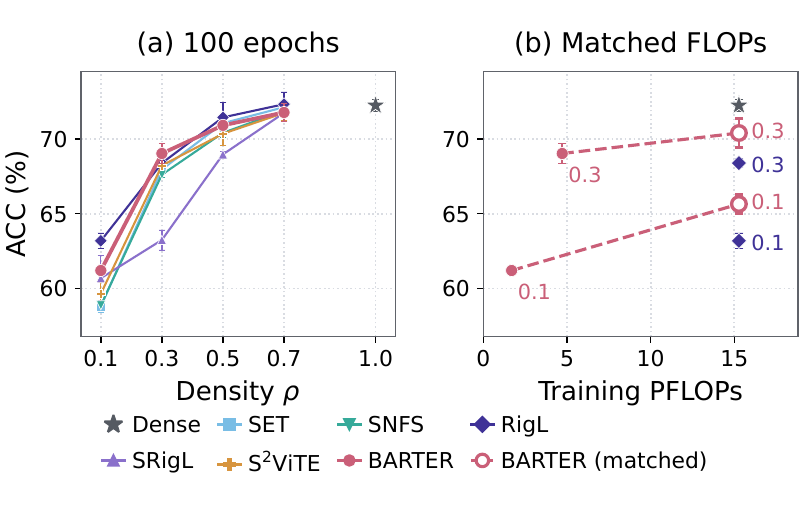}
    \caption{Fine-tuning on CIFAR-100. 
    (a) Results after 100 epochs at sparse densities
    $\rho=0.1,0.3,0.5,0.7$.
    (b) \method{}'s 100-epoch (filled circles) and
    compute-matched (hollow circles) results, compared
    with RigL at the same densities.
    Numeric labels indicate density; Dense is a
    full-density reference.
    Error bars show sample standard deviations.}
    \label{fig:finetuning}
    \vspace{\dimexpr-\intextsep-20pt\relax}
\end{wrapfigure}

At $\rho=0.1$, \method{} reduces estimated training FLOPs
by $89.4\%$ on CCT and $88.6\%$ on ViT relative to Dense
and all compared DST baselines.
The exported models contain $0.480$M and $0.373$M parameters, compared with $3.783$M and $2.762$M for Dense, SET, SNFS, RigL, and SRigL. S$^2$ViTE reaches the same exported parameter counts as \method{}, but uses full-size tensors during training in our implementation. \method{} uses compact tensors during both training and deployment. Complete numerical results and measurement details appear in Appendices~\ref{app:efficiency_results} and~\ref{app:efficiency}, respectively.

\subsection{Fine-Tuning Performance and Efficiency}
\label{sec:finetuning}

\paragraph{Standard fine-tuning.}
We fine-tune the Tiny-ImageNet-pretrained ViT on CIFAR-100.
Figure~\ref{fig:finetuning}(a) compares the results after
100 epochs.
{\bfseries\boldmath
At $\rho=0.3$, \method{} achieves the highest mean ACC
and macro-F1 among sparse methods, at $69.02\%$ and
$68.81\%$, while using $69.2\%$ fewer training FLOPs
than Dense and all compared DST baselines.}
At $\rho=0.1$, the FLOP reduction reaches $89.0\%$,
with ACC $1.98$ percentage points below RigL.

\paragraph{Fine-tuning under a matched FLOP budget.}
We then use the saved compute to train longer at
$\rho=0.1$ and $0.3$.
Matching the baselines' 100-epoch budget of approximately $15.276$ PFLOPs allows \method{} to train for $906.17$ and $324.28$ epochs in total, respectively. ACC rises from $61.21\%$ to $65.65\%$ at $\rho=0.1$
and from $69.02\%$ to $70.39\%$ at $\rho=0.3$;
macro-F1 reaches $65.33\%$ and $70.18\%$. {\bfseries\boldmath
Under the matched compute budget, \method{} exceeds
RigL, the strongest sparse baseline at both densities,
by $2.46$ and $2.01$ percentage points in ACC.}
Figure~\ref{fig:finetuning}(b) shows this comparison.
Complete results, training lengths, and cumulative FLOPs
are in Appendix~\ref{app:finetuning_results}.

\subsection{On-Device Training}
\label{sec:on_device}
\begin{wraptable}{r}{0.40\textwidth}
    \vspace{-\intextsep}
    \centering
    \captionsetup{font=small,skip=4pt}
    \caption{CCT training on iPhone 13 Pro Max.
    RigL and \method{} use $\rho=0.3$.}
    \label{tab:on_device}
    \small
    \setlength{\tabcolsep}{4pt}
    \renewcommand{\arraystretch}{1.08}
    \begin{tabular*}{\linewidth}{@{\extracolsep{\fill}}lrr@{}}
    \toprule
    & \multicolumn{2}{c}{Training time} \\
    \cmidrule(lr){2-3}
    Method & ms/step & s/epoch \\
    \midrule
    Dense & 4814.77 & 1726.77 \\
    RigL & 4927.10 & 1699.36 \\
    \textbf{\method{}} & \textbf{1804.57} & \textbf{610.33} \\
    \bottomrule
    \end{tabular*}
    \vspace{-\intextsep}
\end{wraptable}

We benchmark CCT training on CIFAR-100 using an
iPhone 13 Pro Max, comparing Dense, RigL, and \method{}.
Both sparse methods use $\rho=0.3$.
Table~\ref{tab:on_device} reports training time per
optimizer step and the directly measured time for one epoch.
{\bfseries\boldmath
\method{} completes the measured epoch in $610.33$ seconds,
with speedups of $2.83\times$ over Dense and $2.78\times$
over RigL.}
Its per-step speedups are $2.67\times$ and $2.73\times$,
respectively.
These results show that compact training tensors also
reduce execution time on the phone.
The runtime and timing protocol are detailed in
Appendix~\ref{app:on_device_measurement}.

\subsection{Ablation Studies}
\label{sec:ablation}

We examine capacity reallocation and its selection rules on CCT/CIFAR-100 at $\rho=0.5$. Static trains the same initial compact architecture as \method{} but disables capacity reallocation. Random receiver randomizes the receiving layer and clone parent; Random donor randomizes the unit to prune; Random pair applies both randomizations (Appendix~\ref{app:baseline_ablation_settings}).

\begin{wrapfigure}{r}{0.5\textwidth}
    \vspace{\dimexpr-\intextsep-10pt\relax}
    \centering
    \captionsetup{font=small,skip=4pt}
    \includegraphics[width=\linewidth]
        {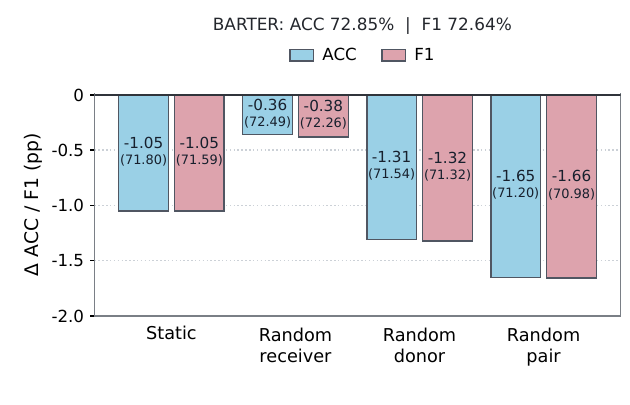}
    \caption{Ablations of capacity reallocation and selection on CCT/CIFAR-100 at $\rho=0.5$. Bars show ACC and macro-F1 differences (variant minus \method{}) in percentage points, with variant scores (\%) in parentheses. Values are means over three seeds per method.}
    \label{fig:selection_ablation}
    \vspace{\dimexpr-\intextsep-3pt\relax}
\end{wrapfigure}

Figure~\ref{fig:selection_ablation} shows that {\bfseries\boldmath \method{} achieves the highest mean ACC and macro-F1, exceeding Static by $1.05$ percentage points in both metrics.} Compared with Random pair, the gains are $1.65$ and $1.66$ percentage points, respectively. Among the single-sided variants, Random donor incurs a larger mean performance drop than Random receiver. \method{} also has lower standard deviations in both metrics than all three randomized variants. Full numerical results are in Appendix~\ref{app:ablation_results}. We extend the comparison with Static to CCT and ViT across all four densities (Appendix~\ref{app:static_comparison}). \method{} achieves higher mean ACC and macro-F1 in every setting, with modest increases in measured training-segment time and similar peak allocated GPU memory.

\section{Conclusion}
\label{sec:conclusion}

We presented \method{}, a structured DST framework that reallocates capacity under fixed global budgets using active-unit statistics while maintaining compact dense execution. Experiments show competitive accuracy, lower peak allocated GPU memory, and speedups in GPU and smartphone training benchmarks. Matched-compute fine-tuning further improves accuracy at low densities, surpassing the strongest sparse baseline. The resulting models require no additional pruning for deployment.

\noindent\textbf{Broader Impact and Future Work.} \method{} supports resource-efficient training without specialized sparse kernels. Future work will explore more effective capacity allocation and extend compact training to environments with changing resource constraints.

\clearpage

\bibliography{iclr2027_conference}
\bibliographystyle{iclr2027_conference}

\clearpage
\appendix
\section{Algorithm and Implementation Details}
\label{app:implementation}

\subsection{Complete Training Algorithm}
\label{app:algorithm}

Algorithm~\ref{alg:barter_complete} summarizes training. Density $\rho$ applies to the attention and FFN weight matrices; embeddings, normalization layers, and the classifier remain outside the pruning pool. Table~\ref{tab:app_capacity_budgets} lists the integer head and channel budgets. We distribute each budget as evenly as possible across layers, assign remainders randomly, and randomly select the retained units within each layer. Their parameters form the initial compact model.

\begin{algorithm}[t]
\caption{Training with \method{}}
\label{alg:barter_complete}
\small
\begin{algorithmic}[1]
\Require Initial model, density $\rho$, training data, and $T$ optimizer steps
\State Construct the compact model with global budgets $H,C$
\State Initialize Adam, zero-valued score EMAs, and the update schedules
\For{$t=1,\ldots,T$}
    \State Run forward and backward passes on an augmented minibatch
    \State Update active-unit EMAs while score collection is enabled
    \State Update model parameters using the task's optimizer and learning-rate schedule
    \For{each unit type scheduled for reallocation at $t$}
        \State Select the highest-scoring eligible receiving layer $r$
        \If{$r$ exists and an eligible donor exists outside $r$}
            \State Select the donor with minimum pruning score and the parent in $r$ with maximum growth score
            \State Clone the parent in $r$, split its outgoing weights, and prune the donor
            \State Resize parameters and Adam states; update scores and cooldowns
        \EndIf
    \EndFor
\EndFor
\State \Return the final model and the checkpoint selected by validation ACC
\end{algorithmic}
\end{algorithm}

\paragraph{Selection and scheduling.}
We use the selection rules in Section~\ref{sec:active_reallocation}, with EMA decay $\beta=0.99$. A receiving layer must contain an active unit and remain below its full-density width. Donors come from other layers, with at least one FFN channel retained per layer. Reallocation takes place between $0.1T$ and $0.4T$, after optimizer updates. FFN proposals are evenly spaced over this window; head proposals use interval midpoints. Clone parents and their copies are excluded from donor selection for two typical FFN proposal intervals or one head interval. Score collection stops after each unit type's final scheduled proposal.

\paragraph{Reallocation budget.}
For the main experiments, we use a limited number of whole-unit transfers to evaluate whether capacity reallocation can improve classification performance in compact models. We scale the scheduled head and FFN-channel transfer counts with their active-unit budgets using reference ratios of $2/34$ and $60/1067$, respectively, corresponding to the CCT configuration at $\rho=0.3$. Counts are rounded to the nearest integer, with at least one transfer scheduled for each feasible unit type. All scheduled transfers were completed in these runs. Table~\ref{tab:app_capacity_budgets} reports the executed transfer counts alongside the global budgets and initial per-layer counts. ViT uses the same reference settings for training from scratch and fine-tuning.

Whole-unit transfers can change a substantial fraction of an individual layer's capacity. At $\rho=0.1$, CCT and ViT initially contain only one or two heads per layer, so one head represents $50\%$ or $100\%$ of a layer's initial head count. FFN channels allow finer adjustments, with 20 to 142 channel transfers per run in the main-experiment settings. Appendix~\ref{app:reallocation_sensitivity} examines how increasing these transfer counts affects classification performance and estimated incremental topology-update cost.

\begin{table}[t]
\centering
\small
\caption{Global capacity budgets and reference transfer counts for the main experiments. Initial per-layer counts are shown in parentheses. All scheduled head and FFN-channel transfers, $K_H$ and $K_C$, were executed in these runs.}
\label{tab:app_capacity_budgets}
\begin{tabular}{@{}lcrrrr@{}}
\toprule
Model & $\rho$ & Heads $H$ & FFN channels $C$ & $K_H$ & $K_C$ \\
\midrule
CCT & $0.1$ & 11 (1--2) & 371 (53) & 1 & 21 \\
 & $0.3$ & 34 (4--5) & 1,067 (152--153) & 2 & 60 \\
 & $0.5$ & 56 (8) & 1,795 (256--257) & 3 & 101 \\
 & $0.7$ & 78 (11--12) & 2,524 (360--361) & 5 & 142 \\
\midrule
ViT & $0.1$ & 11 (1--2) & 359 (39--40) & 1 & 20 \\
 & $0.3$ & 33 (3--4) & 1,033 (114--115) & 2 & 58 \\
 & $0.5$ & 54 (6) & 1,739 (193--194) & 3 & 98 \\
 & $0.7$ & 76 (8--9) & 2,413 (268--269) & 4 & 136 \\
\bottomrule
\end{tabular}
\end{table}

\subsection{Compact Model Implementation}
\label{app:compact_implementation}

Attention projections store only active head blocks, and FFN matrices store only active channel rows and columns. The residual width and per-head dimension remain fixed. Cloning copies the parent's incoming weights and biases, then splits its outgoing weights as in Eq.~(\ref{eq:clone_split}), using Gaussian perturbations with standard deviation $0.02\max(\operatorname{RMS}(v),10^{-8})$ for outgoing block $v$. The receiver is expanded before the donor is pruned; each completed transfer preserves the global head and channel budgets. Zero-head attention skips attention products but retains the output-projection bias and residual connection.

Adam states follow the parameter resizing: pruned slices are removed, unchanged slices are retained, and first- and second-moment slices for both the parent and clone are reset to zero. Optimizer step counters are preserved. The parent and clone each inherit half of the parent's previous pruning and growth EMAs. Persistent unit IDs track cooldowns across resizing. Training and evaluation use the resulting compact tensors directly.

\section{Experimental Details}
\label{app:experimental_details}

\subsection{Datasets and Model Configurations}
\label{app:datasets_models}

For CIFAR-100, we use 45,000 images for training, 5,000 for validation, and the official 10,000-image test set. CCT uses a seeded random split; ViT fine-tuning uses a class-stratified split. For Tiny-ImageNet, we use a class-stratified 90,000/10,000 split of the official training set and evaluate on the official labeled validation set. Within each task and seed, all methods share the same split. Table~\ref{tab:app_model_configurations} lists the architectures. CCT uses a $3\times3$ convolutional tokenizer followed by stride-2 max pooling; ViT uses $8\times8$ patches. Both use learned positional embeddings. Fine-tuning resizes CIFAR-100 images to $64\times64$ by bicubic interpolation and replaces ViT's classifier with a 100-class head.

\begin{table}[t]
\centering
\small
\caption{Full-density model configurations. Fine-tuning retains the ViT backbone.}
\label{tab:app_model_configurations}
\begin{tabular}{@{}lcc@{}}
\toprule
Setting & CCT & ViT \\
\midrule
Input resolution & $32\times32$ & $64\times64$ \\
Transformer layers & 7 & 9 \\
Embedding dimension & 256 & 192 \\
Heads per layer & 16 & 12 \\
Head dimension & 16 & 16 \\
FFN hidden dimension & 512 & 384 \\
Tokens & 256 & $64+1$ \\
Pooling & Attention & Class token \\
\bottomrule
\end{tabular}
\end{table}

We evaluate $\rho\in\{0.1,0.3,0.5,0.7\}$, with Dense as a full-density reference. Within each task, density, and seed, S$^2$ViTE and \method{} share the initial structured allocation. This allocation also sets the active-parameter target for the weight-sparse baselines. For training from scratch, Static uses the same initial compact architecture as \method{}. Baseline and ablation settings are given in Appendix~\ref{app:baseline_ablation_settings}.

\subsection{Training Settings}
\label{app:training_settings}

\paragraph{Optimization.}
Table~\ref{tab:training_hyperparameters} summarizes the training settings. All methods use label-smoothed cross-entropy and Adam with explicitly implemented decoupled weight decay, applied after the Adam update for CCT and before it for ViT training and fine-tuning. Decay includes biases and normalization parameters, and is restricted to active entries in masked implementations. Learning rates use linear warmup followed by cosine decay~\citep{sgdr}. All tasks use FP32 without automatic mixed precision; ViT training and fine-tuning also disable TF32.

\begin{table}[t]
\centering
\small
\caption{Training settings. Fine-tuning epochs refer to the initial phase; warmup is a fraction of optimizer steps.}
\label{tab:training_hyperparameters}
\begin{tabular}{@{}lccc@{}}
\toprule
Setting & CCT & ViT & ViT fine-tuning \\
\midrule
Epochs & 200 & 100 & 100 \\
Batch size & 128 & 256 & 256 \\
Steps per epoch & 351 & 1,054 & 527 \\
Peak learning rate & $3\times10^{-4}$ & $2\times10^{-3}$ & $10^{-3}$ \\
Minimum learning rate & $0$ & $10^{-6}$ & $10^{-6}$ \\
Weight decay & $0.05$ & $0.05$ & $0.05$ \\
Warmup fraction & $0.05$ & $0.10$ & $0.10$ \\
Label smoothing & $0.1$ & $0.1$ & $0.1$ \\
Gradient clipping norm & $1.0$ & None & None \\
Attention dropout & $0.1$ & $0$ & $0$ \\
Maximum stochastic depth & $0.1$ & $0.1$ & $0.1$ \\
Mixup $\alpha$ & $0.2$ & $1.0$ & $1.0$ \\
CutMix $\alpha$ & Disabled & $1.0$ & $1.0$ \\
Batch mixing probability & $1.0$ & $0.5$ & $0.5$ \\
\bottomrule
\end{tabular}
\end{table}

One CCT epoch traverses the training set once. ViT training and fine-tuning use three shuffled passes per epoch, dropping the final incomplete minibatch; their 100-epoch schedules therefore correspond to approximately 300 data passes. Evaluation uses batch size 256. Embedding, FFN, and projection dropout are zero, while stochastic-depth rates increase linearly across layers.

\paragraph{Data augmentation.}
All tasks use random cropping with four-pixel padding and horizontal flipping. CCT additionally uses RandAugment~\citep{randaugment} with two operations and magnitude 9, together with Mixup~\citep{mixup}. ViT training and fine-tuning use ImageNet AutoAugment~\citep{autoaugment} and random erasing with probability $0.25$. When batch mixing is enabled, ViT chooses Mixup or CutMix~\citep{cutmix} with equal probability. Fine-tuning retains the source model's Tiny-ImageNet normalization.

\paragraph{Fine-tuning protocol.}
For each seed, all methods start from the same validation-selected dense Tiny-ImageNet checkpoint, with a newly initialized classifier and optimizer. Sparse methods inherit pretrained weights at their selected positions. After the initial 100 epochs, \method{} at $\rho=0.1$ and $0.3$ continues from the endpoint checkpoint with its topology fixed and score collection disabled. Adam moments and step counters are retained, while a separate warmup/cosine schedule restarts over the remaining compute budget using the same learning-rate settings. Training stops at the baselines' cumulative forward-plus-backward MAC budget, allowing a final partial epoch. The counting protocol is given in Appendix~\ref{app:compute_model_size}.



\subsection{Baseline and Ablation Settings}
\label{app:baseline_ablation_settings}

\paragraph{Baseline implementations.}
We apply the official implementations of SET~\citep{set}, SNFS~\citep{snfs}, RigL~\citep{rigl}, SRigL~\citep{srigl}, and S$^2$ViTE~\citep{s2vite} to the CCT and ViT configurations used in our experiments, following the common training recipes in Appendix~\ref{app:training_settings}. We benchmark all methods using the GPU timing and memory protocol in Appendix~\ref{app:gpu_measurement}.

\paragraph{Static compact baseline.}
We evaluate Static on CCT/CIFAR-100 and ViT/Tiny-ImageNet at $\rho\in\{0.1,0.3,0.5,0.7\}$. Static follows the same training recipe as \method{} but keeps its initial attention-head and FFN-channel allocation fixed throughout training. All retained weights remain trainable, and forward and backward computation use compact tensors. Static uses the same efficiency-measurement protocol as \method{} (Appendix~\ref{app:efficiency}). Appendix~\ref{app:static_comparison} reports the complete classification and efficiency results.

\paragraph{Selection ablations.}
We evaluate the three randomized variants on CCT/CIFAR-100 at $\rho=0.5$, retaining the original model configuration, initial capacity budget, training recipe, and transfer-attempt schedule. Random receiver samples uniformly from layers with at least one active unit and room to grow, then samples a clone parent uniformly from the active units in that layer. Random donor samples uniformly from all eligible active units outside the receiving layer, subject to the same donor cooldown and minimum-width constraints as \method{}. Random pair applies both randomizations. The remaining selection operations use the original scoring rules; score tracking, cloning, and optimizer-state handling are unchanged.

\section{Additional Results}
\label{app:additional_results}

\subsection{Complete Efficiency Results}
\label{app:efficiency_results}

Tables~\ref{tab:efficiency_cct_timing} and~\ref{tab:efficiency_vit_timing} report GPU training time per epoch, peak allocated memory, and schedule-weighted step-time estimates for \method{}, Dense, and the five DST baselines. Tables~\ref{tab:efficiency_cct_compute} and~\ref{tab:efficiency_vit_compute} give estimated training FLOPs and exported parameter counts. The measurement procedures are described in Appendix~\ref{app:efficiency}; comparisons with Static are reported in Appendix~\ref{app:static_comparison}.

\begin{figure}[t]
    \centering
    \includegraphics[width=0.85\textwidth]
        {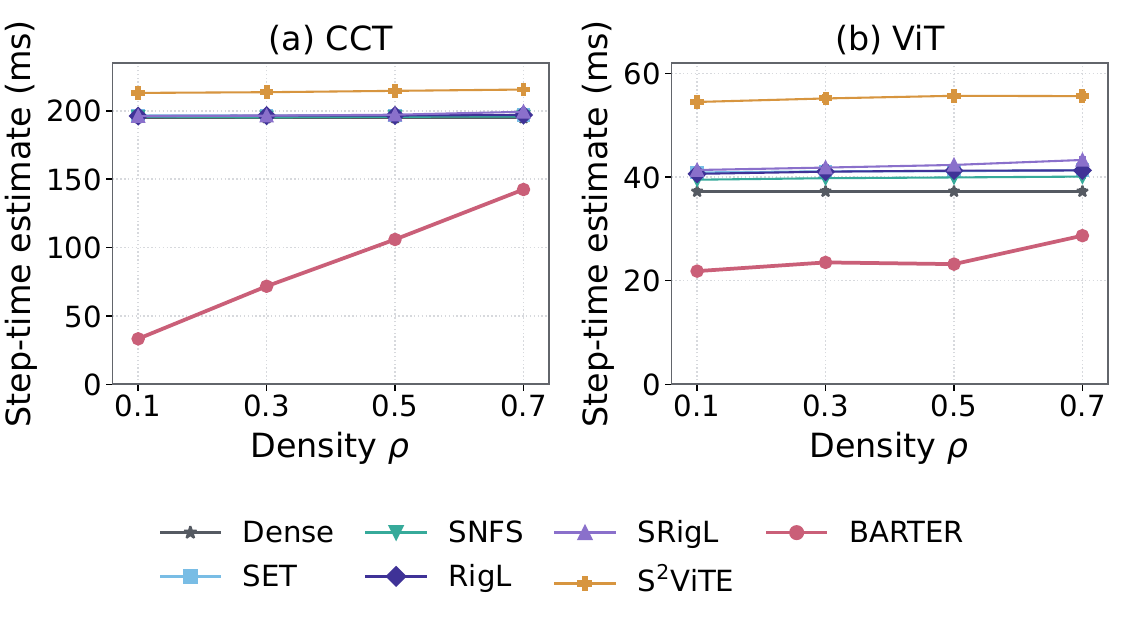}
    \caption{Schedule-weighted training step-time estimates
    for CCT on CIFAR-100 and ViT on Tiny-ImageNet.
    Error bars show sample standard deviations.}
    \label{fig:appendix_step_time}
\end{figure}

\paragraph{Training step time.}
Figure~\ref{fig:appendix_step_time} shows that \method{}
has the lowest mean step-time estimate on both tasks
at every tested density.
At $\rho=0.5$, its estimates are $105.94$\,ms for CCT
and $23.19$\,ms for ViT, compared with $196.59$\,ms
and $41.23$\,ms for RigL.
These correspond to $1.86\times$ and $1.78\times$ speedups,
respectively.

\begin{table}[t]
\centering
\caption{GPU training efficiency for CCT on CIFAR-100.
Values are mean $\pm$ sample standard deviation.
Dense is a full-density reference repeated across columns.}
\label{tab:efficiency_cct_timing}
\begingroup
\small
\setlength{\tabcolsep}{4pt}
\renewcommand{\arraystretch}{1.08}
\begin{tabular*}{\textwidth}{@{\extracolsep{\fill}}lcccc@{}}
\toprule
Method & $\rho=0.1$ & $\rho=0.3$ & $\rho=0.5$ & $\rho=0.7$ \\
\midrule
\multicolumn{5}{l}{\textit{Measured training time (s/epoch)}} \\
Dense & $68.45\pm0.01$ & $68.45\pm0.01$ & $68.45\pm0.01$ & $68.45\pm0.01$ \\
SET & $69.12\pm0.04$ & $69.05\pm0.03$ & $69.11\pm0.02$ & $69.18\pm0.05$ \\
SNFS & $68.73\pm0.03$ & $68.65\pm0.03$ & $68.72\pm0.05$ & $68.80\pm0.07$ \\
RigL & $68.97\pm0.05$ & $69.05\pm0.10$ & $69.09\pm0.02$ & $69.24\pm0.05$ \\
SRigL & $69.27\pm0.05$ & $69.36\pm0.07$ & $69.90\pm0.02$ & $70.78\pm0.22$ \\
S$^2$ViTE & $74.90\pm0.15$ & $75.03\pm0.05$ & $75.45\pm0.17$ & $75.70\pm0.24$ \\
\method{} & $11.90\pm0.06$ & $25.48\pm0.02$ & $37.78\pm0.02$ & $50.88\pm0.11$ \\
\midrule
\multicolumn{5}{l}{\textit{Peak allocated GPU memory (GiB)}} \\
Dense & $11.489\pm0.000$ & $11.489\pm0.000$ & $11.489\pm0.000$ & $11.489\pm0.000$ \\
SET & $11.492\pm0.000$ & $11.492\pm0.000$ & $11.492\pm0.000$ & $11.492\pm0.000$ \\
SNFS & $11.492\pm0.000$ & $11.492\pm0.000$ & $11.492\pm0.000$ & $11.492\pm0.000$ \\
RigL & $11.493\pm0.000$ & $11.493\pm0.000$ & $11.493\pm0.000$ & $11.493\pm0.000$ \\
SRigL & $11.493\pm0.000$ & $11.493\pm0.000$ & $11.493\pm0.000$ & $11.493\pm0.000$ \\
S$^2$ViTE & $11.770\pm0.000$ & $11.770\pm0.000$ & $11.770\pm0.000$ & $11.770\pm0.000$ \\
\method{} & $2.197\pm0.034$ & $4.280\pm0.011$ & $6.335\pm0.002$ & $8.394\pm0.031$ \\
\midrule
\multicolumn{5}{l}{\textit{Schedule-weighted training step estimate (ms/step)}} \\
Dense & $194.78\pm0.06$ & $194.78\pm0.06$ & $194.78\pm0.06$ & $194.78\pm0.06$ \\
SET & $196.71\pm0.07$ & $196.53\pm0.17$ & $196.71\pm0.13$ & $196.91\pm0.10$ \\
SNFS & $195.60\pm0.08$ & $195.27\pm0.04$ & $195.67\pm0.19$ & $195.84\pm0.13$ \\
RigL & $196.18\pm0.17$ & $196.47\pm0.20$ & $196.59\pm0.17$ & $196.98\pm0.16$ \\
SRigL & $196.12\pm0.14$ & $196.45\pm0.19$ & $197.19\pm0.07$ & $199.32\pm0.42$ \\
S$^2$ViTE & $213.12\pm0.46$ & $213.64\pm0.08$ & $214.63\pm0.48$ & $215.59\pm0.56$ \\
\method{} & $33.27\pm0.21$ & $71.71\pm0.20$ & $105.94\pm0.05$ & $142.48\pm0.29$ \\
\bottomrule
\end{tabular*}
\endgroup
\end{table}

\begin{table}[t]
\centering
\caption{Estimated training computation and exported model size for CCT on CIFAR-100.
All sample standard deviations are zero at the reported precision.
Dense is a full-density reference repeated across columns.}
\label{tab:efficiency_cct_compute}
\begingroup
\small
\setlength{\tabcolsep}{4pt}
\renewcommand{\arraystretch}{1.08}
\begin{tabular*}{\textwidth}{@{\extracolsep{\fill}}lcccc@{}}
\toprule
Method & $\rho=0.1$ & $\rho=0.3$ & $\rho=0.5$ & $\rho=0.7$ \\
\midrule
\multicolumn{5}{l}{\textit{Estimated training GFLOPs/image}} \\
Dense & $7.089840$ & $7.089840$ & $7.089840$ & $7.089840$ \\
SET & $7.089840$ & $7.089840$ & $7.089840$ & $7.089840$ \\
SNFS & $7.089840$ & $7.089840$ & $7.089840$ & $7.089840$ \\
RigL & $7.089840$ & $7.089840$ & $7.089840$ & $7.089840$ \\
SRigL & $7.089840$ & $7.089840$ & $7.089840$ & $7.089840$ \\
S$^2$ViTE & $7.089840$ & $7.089840$ & $7.089840$ & $7.089840$ \\
\method{} & $0.750408$ & $2.165988$ & $3.568980$ & $4.972764$ \\
\midrule
\multicolumn{5}{l}{\textit{Exported parameters (M)}} \\
Dense & $3.7833$ & $3.7833$ & $3.7833$ & $3.7833$ \\
SET & $3.7833$ & $3.7833$ & $3.7833$ & $3.7833$ \\
SNFS & $3.7833$ & $3.7833$ & $3.7833$ & $3.7833$ \\
RigL & $3.7833$ & $3.7833$ & $3.7833$ & $3.7833$ \\
SRigL & $3.7833$ & $3.7833$ & $3.7833$ & $3.7833$ \\
S$^2$ViTE & $0.4802$ & $1.2141$ & $1.9480$ & $2.6824$ \\
\method{} & $0.4802$ & $1.2141$ & $1.9480$ & $2.6824$ \\
\bottomrule
\end{tabular*}
\endgroup
\end{table}

\begin{table}[t]
\centering
\caption{GPU training efficiency for ViT on Tiny-ImageNet.
Values are mean $\pm$ sample standard deviation.
Dense is a full-density reference repeated across columns.}
\label{tab:efficiency_vit_timing}
\begingroup
\small
\setlength{\tabcolsep}{4pt}
\renewcommand{\arraystretch}{1.08}
\begin{tabular*}{\textwidth}{@{\extracolsep{\fill}}lcccc@{}}
\toprule
Method & $\rho=0.1$ & $\rho=0.3$ & $\rho=0.5$ & $\rho=0.7$ \\
\midrule
\multicolumn{5}{l}{\textit{Measured training time (s/epoch)}} \\
Dense & $39.78\pm0.48$ & $39.78\pm0.48$ & $39.78\pm0.48$ & $39.78\pm0.48$ \\
SET & $44.50\pm0.92$ & $44.32\pm0.43$ & $44.35\pm0.34$ & $44.76\pm0.51$ \\
SNFS & $42.31\pm0.19$ & $42.44\pm0.40$ & $42.64\pm0.31$ & $42.89\pm0.26$ \\
RigL & $44.03\pm0.25$ & $44.46\pm0.29$ & $44.46\pm0.54$ & $44.58\pm0.45$ \\
SRigL & $45.04\pm0.39$ & $45.19\pm0.63$ & $46.13\pm0.49$ & $47.32\pm0.40$ \\
S$^2$ViTE & $58.36\pm1.28$ & $58.90\pm0.39$ & $59.96\pm1.63$ & $59.77\pm1.22$ \\
\method{} & $25.40\pm0.55$ & $27.22\pm0.29$ & $27.86\pm1.12$ & $31.41\pm0.07$ \\
\midrule
\multicolumn{5}{l}{\textit{Peak allocated GPU memory (GiB)}} \\
Dense & $1.932\pm0.000$ & $1.932\pm0.000$ & $1.932\pm0.000$ & $1.932\pm0.000$ \\
SET & $1.934\pm0.000$ & $1.934\pm0.000$ & $1.934\pm0.000$ & $1.934\pm0.000$ \\
SNFS & $1.934\pm0.000$ & $1.934\pm0.000$ & $1.934\pm0.000$ & $1.934\pm0.000$ \\
RigL & $1.934\pm0.000$ & $1.934\pm0.000$ & $1.934\pm0.000$ & $1.934\pm0.000$ \\
SRigL & $1.944\pm0.000$ & $1.944\pm0.000$ & $1.944\pm0.000$ & $1.944\pm0.000$ \\
S$^2$ViTE & $2.071\pm0.000$ & $2.071\pm0.000$ & $2.071\pm0.000$ & $2.071\pm0.000$ \\
\method{} & $0.701\pm0.001$ & $1.023\pm0.003$ & $1.319\pm0.005$ & $1.633\pm0.001$ \\
\midrule
\multicolumn{5}{l}{\textit{Schedule-weighted training step estimate (ms/step)}} \\
Dense & $37.20\pm0.45$ & $37.20\pm0.45$ & $37.20\pm0.45$ & $37.20\pm0.45$ \\
SET & $40.88\pm0.20$ & $41.10\pm0.40$ & $41.12\pm0.29$ & $41.31\pm0.41$ \\
SNFS & $39.48\pm0.20$ & $39.77\pm0.09$ & $39.92\pm0.02$ & $40.08\pm0.21$ \\
RigL & $40.61\pm0.36$ & $41.03\pm0.16$ & $41.23\pm0.16$ & $41.27\pm0.31$ \\
SRigL & $41.35\pm0.21$ & $41.82\pm0.42$ & $42.33\pm0.23$ & $43.29\pm0.48$ \\
S$^2$ViTE & $54.48\pm0.84$ & $55.16\pm0.71$ & $55.67\pm1.13$ & $55.61\pm0.83$ \\
\method{} & $21.84\pm0.57$ & $23.52\pm0.33$ & $23.19\pm0.18$ & $28.68\pm0.09$ \\
\bottomrule
\end{tabular*}
\endgroup
\end{table}

\begin{table}[t]
\centering
\caption{Estimated training computation and exported model size for ViT on Tiny-ImageNet.
All sample standard deviations are zero at the reported precision.
Dense is a full-density reference repeated across columns.}
\label{tab:efficiency_vit_compute}
\begingroup
\small
\setlength{\tabcolsep}{4pt}
\renewcommand{\arraystretch}{1.08}
\begin{tabular*}{\textwidth}{@{\extracolsep{\fill}}lcccc@{}}
\toprule
Method & $\rho=0.1$ & $\rho=0.3$ & $\rho=0.5$ & $\rho=0.7$ \\
\midrule
\multicolumn{5}{l}{\textit{Estimated training GFLOPs/image}} \\
Dense & $1.137138$ & $1.137138$ & $1.137138$ & $1.137138$ \\
SET & $1.137138$ & $1.137138$ & $1.137138$ & $1.137138$ \\
SNFS & $1.137138$ & $1.137138$ & $1.137138$ & $1.137138$ \\
RigL & $1.137138$ & $1.137138$ & $1.137138$ & $1.137138$ \\
SRigL & $1.137138$ & $1.137138$ & $1.137138$ & $1.137138$ \\
S$^2$ViTE & $1.137138$ & $1.137138$ & $1.137138$ & $1.137138$ \\
\method{} & $0.129786$ & $0.354006$ & $0.577410$ & $0.801624$ \\
\midrule
\multicolumn{5}{l}{\textit{Exported parameters (M)}} \\
Dense & $2.7619$ & $2.7619$ & $2.7619$ & $2.7619$ \\
SET & $2.7619$ & $2.7619$ & $2.7619$ & $2.7619$ \\
SNFS & $2.7619$ & $2.7619$ & $2.7619$ & $2.7619$ \\
RigL & $2.7619$ & $2.7619$ & $2.7619$ & $2.7619$ \\
SRigL & $2.7619$ & $2.7619$ & $2.7619$ & $2.7619$ \\
S$^2$ViTE & $0.3730$ & $0.9039$ & $1.4347$ & $1.9656$ \\
\method{} & $0.3730$ & $0.9039$ & $1.4347$ & $1.9656$ \\
\bottomrule
\end{tabular*}
\endgroup
\end{table}

\subsection{Complete Fine-Tuning Results}
\label{app:finetuning_results}

Table~\ref{tab:finetuning_standard} reports ACC, macro-F1,
and cumulative training FLOPs for all methods after
100 epochs of fine-tuning on CIFAR-100.
Results are means and sample standard deviations over
seeds 1, 2, and 4.
Training FLOPs include the accumulated forward and
backward MAC counts, with two FLOPs per MAC;
one PFLOP equals $10^{15}$ FLOPs.

\begin{table}[t]
\centering
\small
\caption{Fine-tuning results after 100 epochs. ACC and macro-F1
are percentages (mean $\pm$ sample standard deviation over three seeds).
Compute is cumulative forward-plus-backward training PFLOPs.}
\label{tab:finetuning_standard}
\setlength{\tabcolsep}{7pt}
\begin{tabular}{lcccc}
\toprule
Method & Density & ACC (\%) & F1 (\%) & PFLOPs \\
\midrule
Dense & 1.0 & $72.23\pm0.42$ & $72.01\pm0.45$ & 15.27612 \\
\midrule
SET & 0.1 & $58.81\pm0.42$ & $58.26\pm0.48$ & 15.27612 \\
SNFS & 0.1 & $58.89\pm0.26$ & $58.25\pm0.29$ & 15.27612 \\
RigL & 0.1 & $63.19\pm0.53$ & $62.80\pm0.56$ & 15.27612 \\
SRigL & 0.1 & $60.70\pm1.51$ & $60.27\pm1.55$ & 15.27612 \\
S$^2$ViTE & 0.1 & $59.66\pm0.74$ & $59.01\pm0.74$ & 15.27612 \\
\method{} & 0.1 & $61.21\pm0.12$ & $60.68\pm0.17$ & 1.68580 \\
\midrule
SET & 0.3 & $67.96\pm0.22$ & $67.71\pm0.24$ & 15.27612 \\
SNFS & 0.3 & $67.61\pm0.21$ & $67.36\pm0.24$ & 15.27612 \\
RigL & 0.3 & $68.38\pm0.31$ & $68.10\pm0.34$ & 15.27612 \\
SRigL & 0.3 & $63.25\pm0.67$ & $62.91\pm0.69$ & 15.27612 \\
S$^2$ViTE & 0.3 & $68.22\pm0.28$ & $68.02\pm0.27$ & 15.27612 \\
\method{} & 0.3 & $69.02\pm0.66$ & $68.81\pm0.72$ & 4.71074 \\
\midrule
SET & 0.5 & $71.04\pm0.60$ & $70.82\pm0.62$ & 15.27612 \\
SNFS & 0.5 & $70.40\pm0.33$ & $70.11\pm0.33$ & 15.27612 \\
RigL & 0.5 & $71.43\pm1.01$ & $71.19\pm1.05$ & 15.27612 \\
SRigL & 0.5 & $68.96\pm0.22$ & $68.70\pm0.26$ & 15.27612 \\
S$^2$ViTE & 0.5 & $70.34\pm0.80$ & $70.14\pm0.81$ & 15.27612 \\
\method{} & 0.5 & $70.90\pm0.18$ & $70.69\pm0.19$ & 7.72472 \\
\midrule
SET & 0.7 & $72.11\pm0.42$ & $71.89\pm0.40$ & 15.27612 \\
SNFS & 0.7 & $71.88\pm0.42$ & $71.69\pm0.40$ & 15.27612 \\
RigL & 0.7 & $72.31\pm0.82$ & $72.09\pm0.85$ & 15.27612 \\
SRigL & 0.7 & $71.73\pm0.50$ & $71.53\pm0.49$ & 15.27612 \\
S$^2$ViTE & 0.7 & $71.73\pm0.49$ & $71.52\pm0.44$ & 15.27612 \\
\method{} & 0.7 & $71.76\pm0.60$ & $71.58\pm0.61$ & 10.74966 \\
\bottomrule
\end{tabular}
\end{table}

Table~\ref{tab:finetuning_matched} compares \method{}'s
100-epoch and compute-matched runs with RigL at
$\rho=0.1$ and $0.3$.
The reference budget is $15.27612$ PFLOPs, used by
Dense and every compared DST baseline over 100 epochs.
The two compute-matched runs use $15.27616$ and
$15.27618$ PFLOPs over $906.17$ and $324.28$ total epochs,
respectively.
Relative to 100-epoch \method{}, ACC increases by
$4.44$ and $1.37$ percentage points, and macro-F1 by
$4.65$ and $1.37$ percentage points.
At the matched budget, \method{} has higher mean ACC
and macro-F1 than every compared sparse baseline
at the same density.

\begin{table}[t]
\centering
\small
\caption{Fine-tuning with additional compute-matched training.
RigL is the strongest sparse baseline at both densities.
ACC and macro-F1 are percentages (mean $\pm$ sample standard deviation).}
\label{tab:finetuning_matched}
\setlength{\tabcolsep}{5pt}
\begin{tabular}{lccccc}
\toprule
Method & Density & Epochs & PFLOPs & ACC (\%) & F1 (\%) \\
\midrule
RigL & 0.1 & 100.00 & 15.27612 & $63.19\pm0.53$ & $62.80\pm0.56$ \\
\method{} & 0.1 & 100.00 & 1.68580 & $61.21\pm0.12$ & $60.68\pm0.17$ \\
\method{} & 0.1 & 906.17 & 15.27616 & $65.65\pm0.59$ & $65.33\pm0.62$ \\
\midrule
RigL & 0.3 & 100.00 & 15.27612 & $68.38\pm0.31$ & $68.10\pm0.34$ \\
\method{} & 0.3 & 100.00 & 4.71074 & $69.02\pm0.66$ & $68.81\pm0.72$ \\
\method{} & 0.3 & 324.28 & 15.27618 & $70.39\pm0.98$ & $70.18\pm0.95$ \\
\bottomrule
\end{tabular}
\end{table}

\subsection{Complete Ablation Results}
\label{app:ablation_results}

Table~\ref{tab:ablation_complete} reports the scores underlying Figure~\ref{fig:selection_ablation}. The \method{} reference follows Table~\ref{tab:classification}. Variant definitions are given in Appendix~\ref{app:baseline_ablation_settings}.

\begin{table}[t]
    \centering
    \caption{Ablations on CCT/CIFAR-100 at $\rho=0.5$. ACC and macro-F1 (\%) are reported as mean $\pm$ sample standard deviation over three seeds per method. Best means are \textcolor{barterorange}{\textbf{bold and orange}}.}
    \label{tab:ablation_complete}
    \begingroup
    \small
    \setlength{\tabcolsep}{7pt}
    \renewcommand{\arraystretch}{1.12}
    \begin{tabular}{@{}lcc@{}}
        \toprule
        Method & ACC & F1 \\
        \midrule
        Static          & $71.80_{\pm0.14}$ & $71.59_{\pm0.16}$ \\
        Random receiver & $72.49_{\pm0.29}$ & $72.26_{\pm0.30}$ \\
        Random donor    & $71.54_{\pm0.75}$ & $71.32_{\pm0.75}$ \\
        Random pair     & $71.20_{\pm0.65}$ & $70.98_{\pm0.65}$ \\
        \midrule
        \method{} & \textcolor{barterorange}{$\mathbf{72.85}_{\pm0.16}$} & \textcolor{barterorange}{$\mathbf{72.64}_{\pm0.14}$} \\
        \bottomrule
    \end{tabular}
    \endgroup
\end{table}

\subsection{Comparison with Static Compact Training}
\label{app:static_comparison}

We compare \method{} with Static across all four densities on CCT/CIFAR-100 and ViT/Tiny-ImageNet. Both start from the same compact architecture and use the same training settings and efficiency-measurement protocol; Static keeps its capacity allocation fixed.

\paragraph{Classification performance.} Table~\ref{tab:static_complete} shows that \method{} achieves higher mean ACC and macro-F1 at every density on both tasks. ACC gains range from $0.27$ to $2.97$ percentage points on CCT and from $0.34$ to $0.63$ on ViT. The \method{} results are reproduced from Table~\ref{tab:classification}.

\begin{table}[t]
\centering
\small
\caption{Static compact training versus \method{} on the two from-scratch tasks. ACC and macro-F1 (\%) are mean $\pm$ sample standard deviation over three seeds per method. The higher mean in each comparison is \textcolor{barterorange}{\textbf{bold and orange}}.}
\label{tab:static_complete}
\begingroup
\setlength{\tabcolsep}{5pt}
\renewcommand{\arraystretch}{1.12}
\begin{tabular}{@{}ccccc@{}}
\toprule
& \multicolumn{2}{c}{Static} & \multicolumn{2}{c}{\method{}} \\
\cmidrule(lr){2-3}\cmidrule(lr){4-5}
$\rho$ & ACC & Macro-F1 & ACC & Macro-F1 \\
\midrule
\multicolumn{5}{c}{\textbf{CCT/CIFAR-100}} \\
\addlinespace[2pt]
$0.1$ & $63.16_{\pm0.27}$ & $62.73_{\pm0.25}$ & \textcolor{barterorange}{$\mathbf{66.13}_{\pm0.40}$} & \textcolor{barterorange}{$\mathbf{65.79}_{\pm0.44}$} \\
$0.3$ & $70.43_{\pm0.13}$ & $70.27_{\pm0.15}$ & \textcolor{barterorange}{$\mathbf{71.38}_{\pm0.36}$} & \textcolor{barterorange}{$\mathbf{71.21}_{\pm0.32}$} \\
$0.5$ & $71.80_{\pm0.14}$ & $71.59_{\pm0.16}$ & \textcolor{barterorange}{$\mathbf{72.85}_{\pm0.16}$} & \textcolor{barterorange}{$\mathbf{72.64}_{\pm0.14}$} \\
$0.7$ & $72.99_{\pm0.46}$ & $72.75_{\pm0.48}$ & \textcolor{barterorange}{$\mathbf{73.26}_{\pm0.28}$} & \textcolor{barterorange}{$\mathbf{73.05}_{\pm0.28}$} \\
\midrule
\multicolumn{5}{c}{\textbf{ViT/Tiny-ImageNet}} \\
\addlinespace[2pt]
$0.1$ & $42.15_{\pm0.81}$ & $40.77_{\pm0.93}$ & \textcolor{barterorange}{$\mathbf{42.56}_{\pm0.86}$} & \textcolor{barterorange}{$\mathbf{41.14}_{\pm0.99}$} \\
$0.3$ & $51.08_{\pm0.43}$ & $50.54_{\pm0.42}$ & \textcolor{barterorange}{$\mathbf{51.42}_{\pm0.39}$} & \textcolor{barterorange}{$\mathbf{50.81}_{\pm0.35}$} \\
$0.5$ & $52.22_{\pm0.67}$ & $51.82_{\pm0.69}$ & \textcolor{barterorange}{$\mathbf{52.77}_{\pm0.23}$} & \textcolor{barterorange}{$\mathbf{52.34}_{\pm0.29}$} \\
$0.7$ & $52.37_{\pm0.28}$ & $51.94_{\pm0.31}$ & \textcolor{barterorange}{$\mathbf{53.00}_{\pm0.28}$} & \textcolor{barterorange}{$\mathbf{52.64}_{\pm0.23}$} \\
\bottomrule
\end{tabular}
\endgroup
\end{table}

\paragraph{Training efficiency.} Table~\ref{tab:static_timing} reports the corresponding execution costs. \method{} achieves higher mean ACC and macro-F1 than Static at every tested density on both tasks, with modest increases in measured training-segment time and similar peak allocated GPU memory under the same measurement protocol.

\begin{table}[t]
\centering
\small
\caption{GPU training efficiency of Static and \method{}. Training time is measured over an epoch-length segment; step time is a schedule-weighted estimate. Values are mean $\pm$ sample standard deviation. Both methods use the same hardware, batch size, segment length, and measurement protocol.}
\label{tab:static_timing}
\begingroup
\setlength{\tabcolsep}{5pt}
\renewcommand{\arraystretch}{1.12}
\begin{tabular}{@{}clccc@{}}
\toprule
$\rho$ & Method & \shortstack{Training time\\(s/segment)} & \shortstack{Step time\\(ms/step)} & \shortstack{Peak memory\\(GiB)} \\
\midrule
\multicolumn{5}{c}{\textbf{CCT/CIFAR-100}} \\
\addlinespace[2pt]
$0.1$ & Static & $11.53_{\pm0.03}$ & $32.69_{\pm0.05}$ & $2.222_{\pm0.000}$ \\
$0.1$ & \method{} & $11.90_{\pm0.06}$ & $33.27_{\pm0.21}$ & $2.197_{\pm0.034}$ \\
\addlinespace[2pt]
$0.3$ & Static & $25.18_{\pm0.03}$ & $71.49_{\pm0.02}$ & $4.263_{\pm0.001}$ \\
$0.3$ & \method{} & $25.48_{\pm0.02}$ & $71.71_{\pm0.20}$ & $4.280_{\pm0.011}$ \\
\addlinespace[2pt]
$0.5$ & Static & $36.90_{\pm0.08}$ & $104.92_{\pm0.29}$ & $6.296_{\pm0.000}$ \\
$0.5$ & \method{} & $37.78_{\pm0.02}$ & $105.94_{\pm0.05}$ & $6.335_{\pm0.002}$ \\
\addlinespace[2pt]
$0.7$ & Static & $49.70_{\pm0.06}$ & $141.23_{\pm0.23}$ & $8.353_{\pm0.031}$ \\
$0.7$ & \method{} & $50.88_{\pm0.11}$ & $142.48_{\pm0.29}$ & $8.394_{\pm0.031}$ \\
\midrule
\multicolumn{5}{c}{\textbf{ViT/Tiny-ImageNet}} \\
\addlinespace[2pt]
$0.1$ & Static & $24.49_{\pm0.58}$ & $21.72_{\pm0.07}$ & $0.720_{\pm0.000}$ \\
$0.1$ & \method{} & $25.40_{\pm0.55}$ & $21.84_{\pm0.57}$ & $0.701_{\pm0.001}$ \\
\addlinespace[2pt]
$0.3$ & Static & $25.65_{\pm0.80}$ & $22.73_{\pm0.75}$ & $0.989_{\pm0.000}$ \\
$0.3$ & \method{} & $27.22_{\pm0.29}$ & $23.52_{\pm0.33}$ & $1.023_{\pm0.003}$ \\
\addlinespace[2pt]
$0.5$ & Static & $26.91_{\pm0.30}$ & $22.38_{\pm0.25}$ & $1.252_{\pm0.000}$ \\
$0.5$ & \method{} & $27.86_{\pm1.12}$ & $23.19_{\pm0.18}$ & $1.319_{\pm0.005}$ \\
\addlinespace[2pt]
$0.7$ & Static & $30.53_{\pm0.15}$ & $28.38_{\pm0.38}$ & $1.541_{\pm0.000}$ \\
$0.7$ & \method{} & $31.41_{\pm0.07}$ & $28.68_{\pm0.09}$ & $1.633_{\pm0.001}$ \\
\bottomrule
\end{tabular}
\endgroup
\end{table}

\paragraph{Compute and model size.} Across both tasks and all tested densities, \method{} achieves higher mean ACC and macro-F1 than Static with the same exported parameter counts and nearly identical estimated training FLOPs per image (Table~\ref{tab:static_compute}).

\begin{table}[t]
\centering
\small
\caption{Estimated training compute and exported model size. Training GFLOPs per image are rounded to three decimals; parameter counts use four decimals. Reported standard deviations are zero at these precisions. Static parameter counts use \texttt{last.pt}; \method{} counts use the validation-selected \texttt{best.pt}.}
\label{tab:static_compute}
\begingroup
\setlength{\tabcolsep}{5pt}
\renewcommand{\arraystretch}{1.12}
\begin{tabular}{@{}ccccc@{}}
\toprule
& \multicolumn{2}{c}{Training GFLOPs/image} & \multicolumn{2}{c}{Parameters (M)} \\
\cmidrule(lr){2-3}\cmidrule(lr){4-5}
$\rho$ & Static & \method{} & Static & \method{} \\
\midrule
\multicolumn{5}{c}{\textbf{CCT/CIFAR-100}} \\
\addlinespace[2pt]
$0.1$ & $0.750$ & $0.750$ & $0.4802$ & $0.4802$ \\
$0.3$ & $2.166$ & $2.166$ & $1.2141$ & $1.2141$ \\
$0.5$ & $3.569$ & $3.569$ & $1.9480$ & $1.9480$ \\
$0.7$ & $4.973$ & $4.973$ & $2.6824$ & $2.6824$ \\
\midrule
\multicolumn{5}{c}{\textbf{ViT/Tiny-ImageNet}} \\
\addlinespace[2pt]
$0.1$ & $0.130$ & $0.130$ & $0.3730$ & $0.3730$ \\
$0.3$ & $0.354$ & $0.354$ & $0.9039$ & $0.9039$ \\
$0.5$ & $0.577$ & $0.577$ & $1.4347$ & $1.4347$ \\
$0.7$ & $0.802$ & $0.802$ & $1.9656$ & $1.9656$ \\
\bottomrule
\end{tabular}
\endgroup
\end{table}

\subsection{Increasing the Number of Capacity Transfers}
\label{app:reallocation_sensitivity}

We further examine whether additional capacity transfers improve classification performance on ViT/Tiny-ImageNet at $\rho=0.5$. The $1\times$ reference setting uses three head transfers and 98 FFN-channel transfers, matching the main experiments. We increase both counts by factors of two, four, and six while retaining the same initial capacity allocation, global head and FFN-channel budgets, and 100-epoch training recipe. All settings follow the same reallocation rules within the window from $0.1T$ to $0.4T$, with one unit transferred at each scheduled update.

\paragraph{Classification performance.}
Table~\ref{tab:reallocation_budget_accuracy} shows higher mean ACC and macro-F1 at each larger tested transfer count. Increasing the counts from $1\times$ to $6\times$ raises mean ACC from $52.77\%$ to $53.49\%$ and mean macro-F1 from $52.34\%$ to $53.11\%$. These gains of $0.72$ and $0.77$ percentage points come from additional reallocation under the same active-unit budgets.

\begin{table}[t]
\centering
\small
\caption{Effect of increasing capacity-transfer counts on ViT/Tiny-ImageNet at $\rho=0.5$. ACC and macro-F1 are mean $\pm$ sample standard deviation over seeds 1, 2, and 4. Executed transfer counts and classification scores are taken from the same validation-selected checkpoints. Best means are \textcolor{barterorange}{\textbf{bold and orange}}.}
\label{tab:reallocation_budget_accuracy}
\begingroup
\setlength{\tabcolsep}{5pt}
\renewcommand{\arraystretch}{1.12}
\begin{tabular}{@{}lrrrr@{}}
\toprule
Multiplier & Head transfers & FFN transfers & ACC (\%) & Macro-F1 (\%) \\
\midrule
$1\times$ (reference) & 3 & 98 & $52.77_{\pm0.23}$ & $52.34_{\pm0.29}$ \\
$2\times$ & 6 & 196 & $53.15_{\pm0.89}$ & $52.69_{\pm0.94}$ \\
$4\times$ & 12 & 392 & $53.30_{\pm0.60}$ & $52.92_{\pm0.66}$ \\
$6\times$ & 18 & 588 & \textcolor{barterorange}{$\mathbf{53.49}_{\pm0.48}$} & \textcolor{barterorange}{$\mathbf{53.11}_{\pm0.59}$} \\
\bottomrule
\end{tabular}
\endgroup
\end{table}

\paragraph{Update cost.}
Table~\ref{tab:reallocation_budget_cost} reports schedule-weighted estimates of incremental topology-update cost, following Appendix~\ref{app:gpu_measurement}. From $1\times$ to $6\times$, the amortized increment increases from $0.019$ to $0.117$ ms per optimizer step, corresponding to cumulative estimates of $1.989$ and $12.321$ seconds over the 100-epoch schedule. The accuracy gains at $6\times$ therefore accompany an estimated additional topology-update cost of $10.332$ seconds relative to the reference setting.

\begin{table}[t]
\centering
\small
\caption{Estimated incremental topology-update cost for the settings in Table~\ref{tab:reallocation_budget_accuracy}. Values are mean $\pm$ sample standard deviation across three seeds. Per-step costs are averaged over all optimizer steps; cumulative costs cover the 100-epoch schedule.}
\label{tab:reallocation_budget_cost}
\begingroup
\setlength{\tabcolsep}{8pt}
\renewcommand{\arraystretch}{1.12}
\begin{tabular}{@{}lcc@{}}
\toprule
& \multicolumn{2}{c}{Incremental topology-update cost} \\
\cmidrule(lr){2-3}
Multiplier & ms/step & s/schedule \\
\midrule
$1\times$ (reference) & $0.019_{\pm0.001}$ & $1.989_{\pm0.069}$ \\
$2\times$ & $0.038_{\pm0.001}$ & $4.051_{\pm0.080}$ \\
$4\times$ & $0.077_{\pm0.003}$ & $8.067_{\pm0.362}$ \\
$6\times$ & $0.117_{\pm0.005}$ & $12.321_{\pm0.494}$ \\
\bottomrule
\end{tabular}
\endgroup
\end{table}

\section{Efficiency Measurement Protocol}
\label{app:efficiency}

\subsection{GPU Timing and Memory}
\label{app:gpu_measurement}

We benchmark the validation-selected checkpoints using
their stored tensor shapes.
For each method and density, we replay one epoch of its
training schedule: 351 steps with batch size 128 for CCT,
and 1,054 steps with batch size 256 for ViT.
These segments process 44,928 and 269,824 images,
respectively; a ViT epoch follows the three-pass convention
used in training.
The replays use the schedules at epoch 50 for CCT and
epoch 25 for ViT.

After warmup, we synchronize the GPU and time the full
training segment, synchronizing again before stopping
the timer.
The interval includes data loading, device transfer,
augmentation, forward and backward passes, optimizer steps,
score collection, and scheduled topology updates.
We reset the peak-memory counter after warmup and record
the maximum GPU memory allocated during the same interval.

We estimate average step time separately.
Ordinary training steps are grouped by their score-collection
settings, and collection or topology-update windows are timed
as separate segments.
We weight these timings by their frequency in the training
schedule; an update window uses the nearest measured window
of the same type.
Measurements are aggregated within each seed before reporting
the mean and sample standard deviation across seeds.
Speedups are ratios of the reported mean times.

\paragraph{Incremental topology-update cost.}
For the study in Appendix~\ref{app:reallocation_sensitivity}, we benchmark topology-update windows and ordinary training steps using GPU-resident batches. We subtract the cost of an equal number of ordinary steps with matching score-collection settings from each update window and weight the differences over the training schedule. The ordinary-step reference includes the corresponding score collection and EMA updates. We report the resulting increment averaged over all 105,400 optimizer steps in milliseconds per step and accumulated over the 100-epoch schedule in seconds.

\subsection{Training FLOPs and Exported Parameter Counts}
\label{app:compute_model_size}

Both compute measures count only convolution and matrix-multiplication MACs, with two FLOPs per MAC.

\paragraph{Training from scratch.} Let $M_{\mathrm{fwd}}$ denote forward MACs per image, calculated from the tensor shapes executed during training. We approximate the backward cost as twice the forward cost and report estimated training FLOPs per image as
\begin{equation}
\widehat{F}_{\mathrm{train}} = 6M_{\mathrm{fwd}}.
\label{eq:estimated_training_flops}
\end{equation}

\paragraph{Fine-tuning.} We derive forward and backward MAC counts separately from operator shapes observed during training, reusing counts for identical shapes and execution settings. We accumulate these counts over all completed training steps to obtain $M_{\mathrm{train}}$ and report cumulative training FLOPs as $F_{\mathrm{total}}=2M_{\mathrm{train}}$. The matched-compute experiments use this cumulative budget.

\paragraph{Exported parameter counts.} We report all parameters physically stored in the exported models, in millions. Parameter counts use validation-selected checkpoints; Static's counts in Table~\ref{tab:static_compute} use its final checkpoint. S$^2$ViTE removes inactive units during export, whereas Static and \method{} already store compact tensors. Dense, SET, SNFS, RigL, and SRigL retain full-size tensors in the evaluated exports.

\subsection{On-Device Measurement}
\label{app:on_device_measurement}

We run a native Swift implementation of CCT on an
iPhone 13 Pro Max with iOS 18.7.8, using MPSGraph
on the Metal GPU in a Release build.
Training uses FP32 and an effective batch size of 128,
formed by accumulating gradients over 32 microbatches
of four images each.
For each method, we report one step-timing run and
one epoch-timing run.

Step time is the mean over 100 optimizer steps after
30 warmup steps.
It includes synchronized device execution, gradient
accumulation, and the optimizer update; CPU data
preparation is outside this interval.
Epoch time is measured separately over 351 optimizer steps, covering 44,928 CIFAR-100 images,
and includes CPU data preparation.
Initial graph compilation and prewarming precede
both measurements.
\method{}'s score collection and EMA updates are
included in the measured training time.

\end{document}